\documentclass[review]{elsarticle}
\usepackage{wrapfig}
\usepackage{blindtext}
\usepackage{graphicx}
\usepackage{makeidx}
\usepackage{float}
\usepackage{multirow}
\usepackage{subfig}
\usepackage{adjustbox}
\usepackage{amsmath}
\usepackage[table,xcdraw]{xcolor}
\usepackage{booktabs}
\usepackage[normalem]{ulem}
\useunder{\uline}{\ul}{}
\usepackage{amssymb}
\usepackage{pifont}
\newcommand{\xmark}{\ding{55}}%
\usepackage{bbm}
\usepackage{lipsum}
\usepackage{appendix}
\usepackage{cleveref}
\journal{Medical Image Analysis}

\usepackage{numcompress}\biboptions{authoryear}

\begin{document}

\begin{frontmatter}


\title{Fully Convolutional Multi-scale Residual DenseNets for Cardiac Segmentation and Automated Cardiac Diagnosis using Ensemble of Classifiers}

%
%
%

\author[mymainaddress]{Mahendra Khened}
\author[mymainaddress]{Varghese Alex}
\author[mymainaddress]{Ganapathy Krishnamurthi\corref{mycorrespondingauthor}}
\cortext[mycorrespondingauthor]{Corresponding author}
\ead{gankrish@iitm.ac.in}
\address[mymainaddress]{Department of Engineering Design\\ Indian Institute of Technology Madras, Chennai, India}

\begin{abstract}
Deep fully convolutional neural network (FCN) based architectures have shown great potential in medical image segmentation. However, such architectures usually have millions of parameters and inadequate number of training samples leading to over-fitting and poor generalization. In this paper, we present a novel highly parameter and memory efficient FCN based architecture for medical image analysis. We propose a novel up-sampling path which incorporates long skip and short-cut connections to overcome the feature map explosion in FCN like architectures. In order to processes the input images at multiple scales and view points simultaneously, we propose to incorporate Inception module's parallel structures. We also propose a novel dual loss function whose weighting scheme allows to combine advantages of cross-entropy and dice loss. We have validated our proposed network architecture on two publicly available datasets, namely: (i) Automated Cardiac Disease Diagnosis Challenge (ACDC-2017), (ii) Left Ventricular Segmentation Challenge (LV-2011). Our approach in ACDC-2017 challenge stands second place for segmentation and first place in automated cardiac disease diagnosis tasks with an accuracy of 100\%. In the LV-2011 challenge our approach attained 0.74 Jaccard index, which is so far the highest published result in fully automated algorithms. From the segmentation we extracted clinically relevant cardiac parameters and hand-crafted features which reflected the clinical diagnostic analysis to train an ensemble system for cardiac disease classification. Our approach combined both cardiac segmentation and disease diagnosis into a fully automated framework which is computational efficient and hence has the potential to be incorporated in computer-aided diagnosis (CAD) tools for clinical application.     
\end{abstract}


\end{frontmatter}


\section{Introduction}
\label{sec:1}
Cardiac cine Magnetic Resonance (MR) Imaging is primarily used for assessment of cardiac function and diagnosis of Cardiovascular diseases (CVDs). Estimation of clinical parameters such as ejection fraction, ventricular volumes, stroke volume and myocardial mass from cardiac MRI is considered as gold standard. Delineating important organs and structures from volumetric medical images, such as MR and computed tomography (CT) images, is usually considered as the primary step for estimating clinical parameters, disease diagnosis, prediction of prognosis and surgical planning.  
In a clinical setup, a radiologist delineates the region of interest from the surrounding tissues/ organs by manually drawing contours encompassing the structure of interest. However, this approach becomes infeasible in a hospital with high footprint as it is time-consuming, tedious and also introduces intra and inter-rater variability (\cite{petitjean2011review, miller2013quantification,tavakoli2013survey,suinesiaputra2014collaborative}). Hence, a fully automatic method for segmentation and clinical diagnosis is desirable. 

Segmentation of Left Ventricular (LV) Endocardium and Epicardium as well as Right Ventricular (RV) Endocardium from 4D-cine (3D+Time) MR datasets has received significant research attention over past few years and several grand challenges (\cite{radau2009evaluation,suinesiaputra2014collaborative, petitjean2015right, Kaggle2016DSB, ACDC2017}) have been organized for advancing the state of art methods in (semi-)/automated cardiac segmentation. These challenges usually provide expert-ground truth contours and provide set of evaluation metrics to benchmark various approaches.
In this paper, we present a novel 2D Fully Convolutional Neural Network (FCN) architecture for medical image segmentation which is highly parameter and memory efficient.  We also develop a fully automated framework which incorporates cardiac structures segmentation and cardiac disease diagnosis. Our contributions are summarized as follows:

\begin{itemize}
\item We employed conventional computer-vision techniques like Circular Hough Transform and Fourier based analysis as a pre-processing step to localize the region of interest (ROI). The extracted ROI is used by the proposed network during training and inference time. This combined approach helps in reduction of GPU memory usage, inference time and elimination of False Positives.
\item The proposed network connectivity pattern was based on Densely connected convolutional neural networks (DenseNets) (\cite{huang2016densely}). DenseNet facilitates  multi-path flow for gradients between layers during training by back-propagation and hence does implicit deep-supervision. DenseNets also encourages feature reuse and thus substantially reduces the number of parameters while maintaining good performance, which is ideal in scenarios with limited data. In addition, we incorporated multi-scale processing in the initial layers of the network by performing convolutions on the input with different kernel sizes in parallel paths and later fusing them as in Inception architectures. We propose a novel long skip and short-cut connections in the up-sampling path which is much more computationally and memory efficient when compared to standard skip connections. We introduce a weighting scheme for both cross-entropy and dice loss, and propose a way to combine both the loss functions to yield optimal performance in terms of pixel-wise accuracy and segmentation metrics.
\item For the cardiac disease classification, the predicted segmentation labels were used to estimate relevant clinical and hand-crafted cardiac features. We identified and extracted relevant features based on clinical inputs and Random Forest based feature importance analysis. We developed an ensemble classifier system which processed the features in two-stages for prediction of the cardiac disease.
\item We extensively validated our proposed network on two cardiac segmentation tasks: (i) segmentation of Left  Ventricle (LV),  Right Ventricle (RV) and Myocardium (MYO) from 3D cine cardiac MR images for both End Diastolic (ED) and End Systolic (ES) phase instances and (ii) segmentation of MYO for the whole cardiac frames and slices in 4D cine MR images, by participating in two challenges organized at Statistical Atlases and Computational Modeling of the Heart (STACOM) workshops: (i) Automated Cardiac Disease Diagnosis Challenge (\cite{ACDC2017}), and (ii) Left Ventricular Segmentation Challenge (\cite{suinesiaputra2014collaborative}).
\end{itemize}

We achieved competitive segmentation results to state-of-the-art approaches in both the challenges, demonstrating the effectiveness and generalization capability of the proposed network architecture. Also, for the cardiac disease classification model, our approach gave $100$\% accuracy on the ACDC testing dataset.
A preliminary version of this work was presented at STACOM workshop held in conjunction with Medical Image Computing and Computer Assisted Interventions (MICCAI-2017). In this paper, we have substantially improved the network architecture and methodology used for cardiac disease classification. The main modifications include elaborating proposed methods, analyzing underlying network connectivity pattern, loss function and adding experiments on  LV-2011 challenge datasets. Since, our work is primarily based on CNNs, we focus on the recently published CNN-based algorithms for cardiac segmentation and also provide a comprehensive literature review on the other approaches.

\subsection{Related Work}
\subsubsection{Literature survey on short-axis cardiac cine MR segmentation}
\cite{petitjean2011review,frangi2001three, tavakoli2013survey, peng2016review} provide a comprehensive survey on cardiac segmentation using semi-automated and fully automated approaches. These approaches can be broadly classified as techniques based on:
\begin{enumerate}
\item pixel or image-based classification such as intensity distribution modeling and image thresholding  (\cite{lynch2006automatic, pednekar2006automated, nambakhsh2013left, jolly2011automatic, katouzian2006new, lu2009segmentation, cousty2010segmentation})

\item variational and level sets (\cite{fradkin2008comprehensive, lynch2008segmentation, paragios2003level, ayed2008left})

\item dynamic programming (\cite{pednekar2006automated, uzumcu2006time})

\item graph cuts and image-driven approaches (\cite{boykov2000interactive, lin2006model, cocosco2008automatic})

\item deformable models such as active contours (\cite{kaus2004automated, el2007automated, billet2009cardiac, cordero2011unsupervised, queiros2015fast})

\item cardiac atlases based registration ( \cite{lorenzo2004segmentation, lotjonen2004statistical, bai2015multi})

\item statistical shape and active appearance models (\cite{mitchell2001multistage, van2006spasm, ordas2007statistical, zhu2010segmentation, grosgeorge2011automatic, zhang20104, alba2018automatic})

\item learning-based such as neural networks and combination of various other approaches (\cite{margeta2011layered, eslami2013segmentation, tsadok2013automatic, tran2016fully, avendi2016combined, tan2017convolutional, zotti2017gridnet, patravali20172d, isensee2017automatic, wolterink2017automatic, baumgartner2017exploration})

\end{enumerate}

\subsubsection{Fully Convolutional Neural Networks for Medical Image Segmentation}

In the field of medical image analysis, considerable amount of work has been done in the lines of automating segmentation of various anatomical structures, detection and delineation of lesions and tumors. Non-learning based algorithms such as statistical shape modeling, level sets, active contours, multi-atlas and graphical models have shown promising results on limited dataset, but they usually tend to perform poorly on data originating from a database outside the training data. Some of these techniques heavily relied on engineering hand-crafted features and hence required domain knowledge and expert inputs. Moreover, hand-crafted features have limited representational ability to deal with the large variations in appearance and shapes of anatomical organs. 
In order to overcome this limitations, learning based methods have been explored to seek more powerful features.\\
Convolutional Neural Networks (CNNs) which were first invented by \cite{lecun1998gradient} is currently the technique that produces state of the art results for a variety of computer vision and pattern recognition tasks. Most common ones are image classification (\cite{krizhevsky2012imagenet,simonyan2014very,he2016deep}) and semantic segmentation using fully convolutional networks (FCN) (\cite{shelhamer2017fully}). 
In all these applications, CNNs have demonstrated greater representational and hierarchical learning ability. Recently in medical image analysis, FCN and its popular extensions like U-NET (\cite{ronneberger2015u}) have achieved remarkable success in segmentation of various structures in heart (\cite{dou20163d}), brain lesions (\cite{havaei2017brain,kamnitsas2017efficient,pereira2016brain}), liver lesions (\cite{christ2016automatic,dou20163d,ben2015automated,deng20083d}) from medical volumes. Also with availability of huge amounts of labeled data and increase in the computational capability of general purpose graphics processors units (GPUs), CNN based methods have the potential for application in daily clinical practice.

\section{Material and Methods}
\label{sec:2}
\subsection{Overview}

\begin{figure}
\centering
\includegraphics[width=.5\textwidth,keepaspectratio]{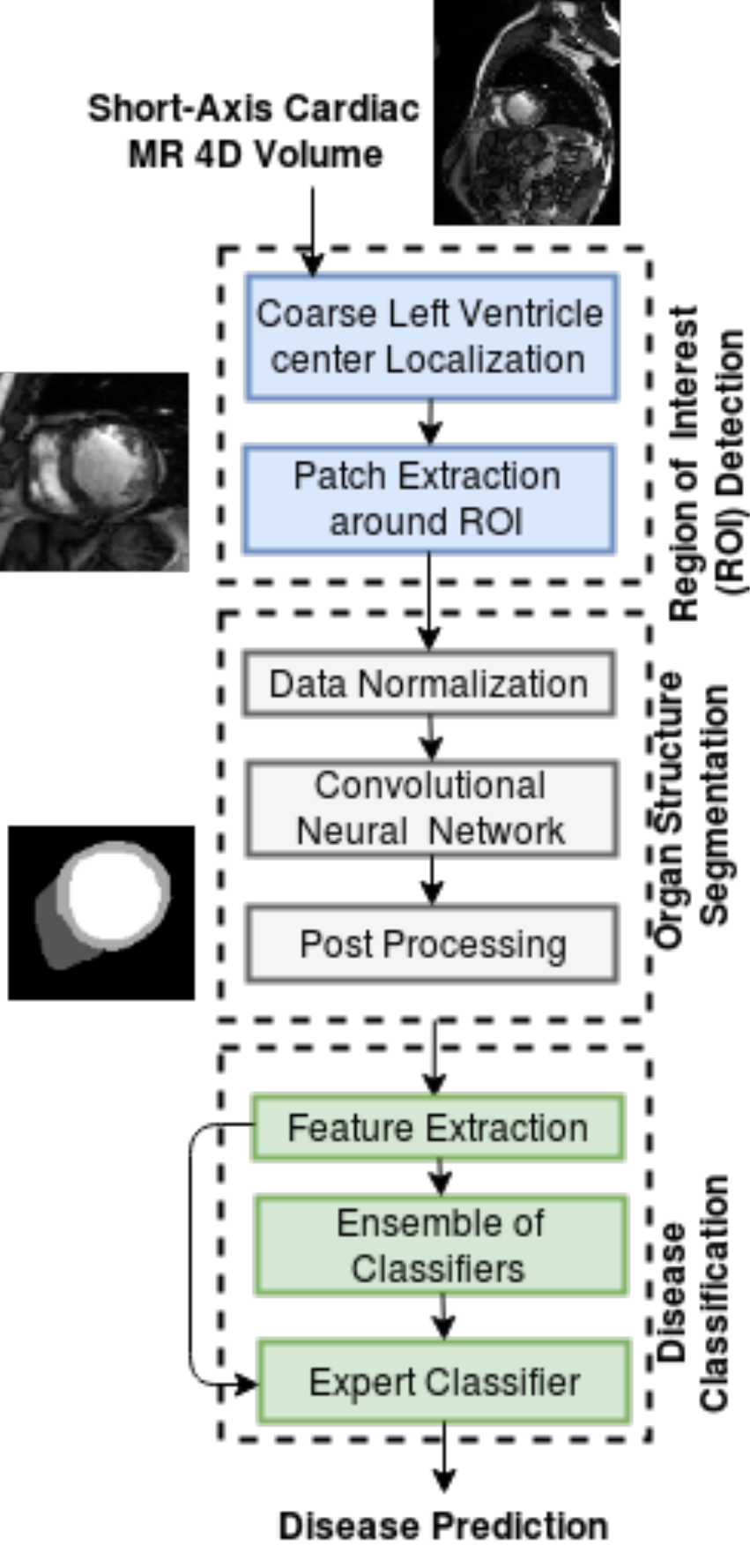}
\caption{ Proposed pipeline for Automated Cardiac Segmentation \& Cardiac Disease diagnosis.}
\label{fig:pipeline}
\end{figure}

Figure \ref{fig:pipeline} illustrates our automated cardiac segmentation and disease diagnosis framework. The pipeline involves: (i) Fourier analysis and Circular Hough-Transform for Region of Interest (ROI) cropping, (ii) proposed network for cardiac structures segmentation and (iii) an ensemble of classifiers for disease diagnosis based on features extracted from the segmentation.

\subsection{Experimental Datasets and Materials}
In this section, we introduce the datasets employed in our experiments. In Section \ref{sec:3}, we shall report our results of comparison study for analyzing the effectiveness of our proposed approach. In Section \ref{sec:4}, we shall comprehensively present the results on these datasets.

\subsubsection{ACDC-2017 Dataset}
The Automated Cardiac Disease Diagnosis challenge dataset comprised of 150 exams of different patients and was divided into 5 evenly distributed subgroups (4 pathological and 1 healthy subject groups) namely- (i) normal- NOR, (ii) patients with previous myocardial infarction- MINF, (iii) patients with dilated cardiomyopathy- DCM, (iv) patients with hypertrophic cardiomyopathy- HCM, (v) patients with abnormal right ventricle- ARV.
Each group was clearly defined according to cardiac physiological parameter, such as the left or right diastolic volume or ejection fraction, the local contraction of the LV, the LV mass and the maximum thickness of the Myocardium. The Cine MR images were acquired in breath hold with a retrospective or prospective gating and with a SSFP sequence in short axis orientation. A series of short axis slices cover the LV from the base to the apex, with a slice thickness of 5-8 mm and an inter-slice gap of 5 or 10 mm. The spatial resolution goes from 1.37 to 1.68 $mm^2$/pixel and 28 to 40 images cover completely or partially the cardiac cycle. For each patient, the weight, height and the diastolic and systolic phase instants were provided. The challenge organizers had evenly divided the patient database based on the pathological condition and was made available in two phases, 100 for training and 50 for testing. The manual annotations of LV, RV and MYO were done by clinical experts at systolic and diastolic phase instances only. The clinical diagnosis and manual annotations were provided for the training set and those of testing set were held out by the challenge organizers for their independent evaluation.
\subsubsection{LV-2011 Dataset}
LV segmentation challenge dataset was made publicly available as part of the STACOM 2011 challenge on automated LV Myocardium segmentation from short-axis cine MRI. The dataset comprised of 200 patients with coronary artery disease and myocardial infarction. The dataset comes with expert-guided semi-automated segmentation contours for the Myocardium. The dataset provided by the organizers was divided into two sets of 100 cases each: training and validation. The training set was provided with expert-guided contours of the Myocardium. 

\subsection{Region of Interest (ROI) detection}
The cardiac MR images of the patient comprises of the heart and the surrounding chest cavity like the lungs and diaphragm. The ROI detection employs Fourier analysis (\cite{lin2006automated}) and Circular Hough Transform (\cite{duda1972use, kunsthart}) to delineate the heart structures from the surrounding tissues. The ROI extraction involves finding the approximate center of the LV and extracting a patch of size  $128\times128$ centered around it. The ROI patch was used for training CNN and also during inference. This approach helped in alleviating the class-imbalance problem associated with labels for heart structures seen in the full sized cardiac MR images. \ref{app:B} gives a detailed overview of ROI detection steps.

\subsection{Normalization}
The 4D cine MR cardiac datasets employed slice-wise normalization of voxel intensities using Eq. (\ref{eq:minmax}).
\begin{equation}
\label{eq:minmax}
X_{norm}=\frac{X - X_{min}}{X_{max}-X_{min}}
\end{equation}
where $X$ is voxel intensity. $X_{min}$ and $X_{max}$ are minimum and maximum of the voxel intensities in a slice respectively.

\subsection{Network Architecture}
The proposed network's connectivity pattern was constructed by taking inspiration from DenseNet (\cite{huang2016densely}) for semantic segmentation (\cite{jegou2017one}). A brief overview on DenseNets, Residual Networks and Inception Architectures is given in \ref{app:C}

\subsubsection{Fully Convolutional Multi-scale Residual DenseNets for Semantic Segmentation}
\begin{figure*}
\centering
\includegraphics[width=0.8\textwidth,keepaspectratio]{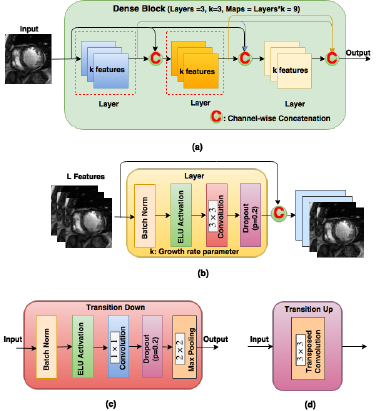}
\caption{The proposed architecture for semantic segmentation using Densely Connected Fully Convolutional Network (DFCN) was made up of several modular blocks which are explained as follows: (a) An example of Dense Block (DB) with 3 Layers. In a DB the input was fed to the first layer to create k new feature maps. These new feature maps are concatenated with the input and fed to the second layer to create another set of k new feature maps. This operation was repeated for 3 times and the final output of the DB was a concatenation of the outputs of all the 3 layers and thus contains 3*k feature maps, (b) A Layer in a DB was a composition of Batch Norm (BN), Exponential Linear Unit (ELU), $3\times 3$ Convolution and a Dropout layer with a drop-out rate of $p=0.2$, (c) A Transition Down (TD) block reduces the spatial resolution of the feature maps as the depth of the network increases. TD block was composed of BN, ELU, $1\times 1$ Convolution, Dropout ($p=0.2$) and $2\times 2$ Max-pooling layers, (d) A Transition Up (TU) block increases the spatial resolution of the feature maps by performing $3\times 3$ Transposed Convolution with a stride of $2$.}
\label{fig:arch}
\end{figure*}

\begin{figure*}
\centering
\includegraphics[width=\textwidth]{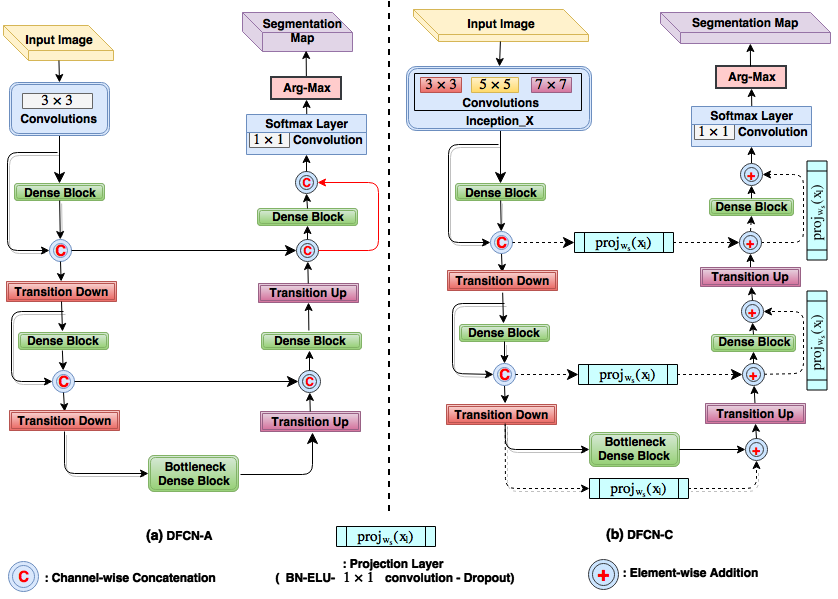}
\caption{The figures shows the modifications introduced to DenseNets to mitigate the feature map explosion when extending to FCN. (a) DFCN-A (\cite{jegou2017one}), (b) The proposed architecture was referred to as DFCN-C. The main modifications include:- (i) replacing the standard copy and concatenation of skip connections from down-sampling path to up-sampling path with a projection layer and an element-wise addition of feature-maps respectively, aiding in reduction of parameters and GPU memory foot-print. (ii) introduction of short-cut connections (residual) in the up-sampling path, (iii) parallel pathways in the initial layer.}
\label{fig:networks}
\end{figure*}

Figure \ref{fig:arch} \& \ref{fig:networks} illustrates the building blocks and the schematic diagram of the proposed network architectures for segmentation respectively. A typical semantic segmentation architecture comprises of a down-sampling path (contracting) and an up-sampling path (expanding). The down-sampling path of the network is similar to the DenseNet architecture and is described in \ref{densenet_overview}. The last layer of the down-sampling path was referred to as bottleneck. The input spatial resolution was recovered in the up-sampling path by transposed convolutions, dense blocks and skip connections coming from the down-sampling path. The up-sampling operation was referred to as Transition-Up (TU). The up-sampled feature maps are added element-wise with skip-connections. The feature maps of the hindmost up-sampling component was convolved with a $1\times1$ convolution layer followed by a soft-max layer to generate the final label map of the segmentation.

In the proposed network architecture several modifications were introduced in its connectivity pattern to further improve on in terms of parameter efficiency, rate of convergence and memory foot-print required. We discuss and compare 3 different architectural variants of Densely Connected Fully Convolutional Networks (DFCN). We refer to the architecture introduced by \cite{jegou2017one} as DFCN-A and other two variants as DFCN-B \& DFCN-C (Fig. \ref{fig:networks}). The following architectural changes were introduced: 
\begin{itemize}

\item The GPU memory foot-print increases with the increase in number of feature maps of larger spatial resolution. In order to mitigate feature map explosion in the up-sampling path, the skip connections from down-sampling path to up-sampling path used element-wise addition operation instead of concatenation operation. In order to match the channel dimensions a projection operation was done on the skip connection path using BN-ELU-$1\times1$-convolution-Dropout. This operation when compared to concatenation of feature maps helps in reduction of the parameters and memory foot-print without affecting the quality of the segmentation output. The proposed projection operation does dimension reduction and also allows complex and learnable interactions of cross channel information (\cite{lin2013network}). Replacing the activation function from rectified linear units (ReLUs) to exponential linear units (ELUs) manifested in faster convergence.

\item In DenseNets, without pooling layers the spatial resolution of feature maps increases with depth and hence leads to memory explosion. So, \cite{jegou2017one} overcame this limitation in the up-sampling path by not concatenating the input to a Dense Block with its output (only exception was at the last dense block, see Fig. \ref{fig:networks} (a)). Hence, the transposed convolution was applied only to the feature maps obtained by the last layer of a dense block and not to all feature maps concatenated so far. However, we observed that by introducing short-cut (residual) connections (\cite{he2016deep}) in the dense blocks of up-sampling path by element-wise addition of dense Blocks input with its output would better aggregate a set of previous transformations rather than completely discarding the inputs of dense blocks. In order to match dimensions of dense blocks input and output a projection operation was done using BN-ELU-$1\times1$-convolution-Dropout. We found that this was effective in addressing the memory limitations exposed by DenseNet based architecture for semantic segmentation. We also observed faster convergence and improved segmentation metrics due to short-cut connections. The DFCN-B refers to architecture got by the above modifications to DFCN-A.

\item DFCN-C (Figure \ref{fig:networks}(b)) is the final architecture that incorporates all the modifications of DFCN-B and additionally its initial layer included parallel CNN branches similar to inception module (\cite{szegedy2015going}). Incorporating multiple kernels of varying receptive fields in each of the parallel paths would help in capturing view-point dependent object variability and learning relations between image structures at multiple-scales (See \ref{app:C} and \ref{fig:inception_x} for more details).
\end{itemize}

\subsubsection{Loss function}
In CNN based techniques, segmentation of volumetric medical images is achieved by performing voxel-wise classification. The network's soft-max layer output gives the posterior probabilities for each class. The anatomical organs or lesions of interest in volumetric medical images are sparsely represented in whole volume leading to class-imbalance, making it difficult to train. In order to address this issue different loss functions such as dice loss (\cite{milletari2016v}) and weighted cross-entropy loss have been introduced.\\
\begin{figure*}
\includegraphics[width=\textwidth]{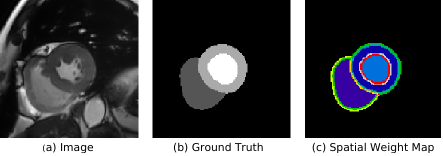}
\caption{The figure shows the spatial weight-map generated from the ground-truth image. The spatial weight map was used with voxel-wise cross-entropy loss. The contour pixels for each class was identified using canny-edge detector with $\sigma = 1$ and was followed by morphological dilation. The colors in spatial weight-map indicate weight distribution based on their relative class frequency.}
\label{fig:wmap}
\end{figure*}
For training the network, a dual loss function which incorporated both cross-entropy and dice loss was proposed. Additionally, two different weighting mechanisms for both the loss functions were introduced. The cross-entropy loss measures voxel-wise error probability between the predicted output class and the target class and accumulates the error across all the voxels. Spatial weight maps generated from the ground-truth images (Fig. \ref{fig:wmap}) were used  for weighting the loss computed at each voxel in the cross-entropy loss Eq. (\ref{eq:cross_entropy_loss}).

Let  $W \,= \, (w_{1}, w_{2}, \cdots, w_{l})$ be the set of learn-able weights, where $w_{l}$ is weight matrix corresponding to the $lth$ layer of the deep network, $p(t_i|x_i; W)$ represent the probability prediction of a voxel $x_i$ after the soft-max function in the last output layer, the spatially weighted cross-entropy loss is formulated as: 
\begin{equation}
\label{eq:cross_entropy_loss}
L_{CE}(X; W) = - \sum_{x_i \in X} w_{map}(x_i) log(p(t_i|x_i; W))
\end{equation}
where $X$ represents the training samples and $t_i$ is the target class label corresponding to voxel $x_i \in X$ and $w_{map}(x_i)$ is the weight estimated at each voxel $x_i$.

Let $L$ be the set of all ground truth classes in the training set. For each ground-truth image, let $N$ be the set of all voxels, $T_l$ be the set of voxels corresponding to each class $l \in L$ and $C_{l}$ be the set of contour voxels corresponding to each class $l \in L$.

\begin{equation}
w_{map}(x_i) = \sum_{l \in L}\frac{|N|*\mathbbm{1}_{T_l}(x_i)}{|T_l|} + \sum_{l \in L}\frac{|N|*\mathbbm{1}_{C_l}(x_i)}{|C_l|}
\end{equation}
where $|.|$ denotes the cardinality of the set and $\mathbbm{1}$ represents the indicator function defined on the subsets of $N$, i.e.  $C_l \subset T_l \subset N$, $\forall \, l \in  L$ 
\begin{equation}
\mathbbm{1}_{T_l}(x_i) :=
\begin{cases}
  1 & x_i \in T_l \\
  0 & x_i \not \in T_l
\end{cases}
\end{equation}

\begin{equation}
\mathbbm{1}_{C_l}(x_i) :=
\begin{cases}
  1 & x_i \in C_l \\
  0 & x_i \not \in C_l
\end{cases}
\end{equation}

For using dice overlap coefficient score as loss-function an approximate value of dice-score was estimated by replacing the predicted label ($p_i$ in Eq. (\ref{eq:dice})) with its posterior probability $p(t_i|x_i; W)$. Since, dice-coefficient needs to be maximized for better segmentation output, the optimization was done to minimize its complement, i.e. $(1-\widehat{DICE})$

For multi-class segmentation, the dice loss was computed using weighted mean of $\widehat{DICE_l}$ for each class $l \in L$. The weights were estimated for every mini-batch instance from the training set. The dice loss for multi-class segmentation problem is given in Eq. (\ref{eq:wdice}):

\begin{equation}
\label{eq:dice_approx}
l_{DICE}(X; W) = \frac{\sum_{x_i \in X}p(t_i|x_i; W)g(x_i) + \epsilon}{\sum_{x_i \in X}(p(t_i|x_i; W)^2 + g(x_i)^2) + \epsilon}
\end{equation}
\begin{equation}
\label{eq:wdice}
L_{DICE} = \frac{\sum_{l \in L} w_l l_{DICE}}{\sum_{l \in L} w_l}
\end{equation}
where $w_l$ is the estimated weight for each class $l \in L$ and $\epsilon$ is a small value added to both numerator and denominator for numerical stability. Let $M$ be the set of pixels in the mini-batch, $M_l$ be the set of pixels corresponding to each class $l \in L$ and $M_l \subset M$, then the weight estimate for the current mini-batch is given by: 
\begin{equation}
w_l = \frac{|M|}{|M_l|} \quad  \forall l \in L
\end{equation}
The parameters of the network were optimized to minimize both the loss functions in tandem. In addition, an $L2$ weight-decay penalty was added to the loss function as regularizer. The total loss function is given in Eq. (\ref{eq:loss}).
\begin{equation}
\label{eq:loss}
LOSS = \lambda (L_{CE}) + \gamma (1-L_{DICE}) + \eta ||W||^2
\end{equation}
where $\lambda$, $\gamma$ and $\eta$ are weights to individual losses. The $L2$ loss decay factor was set to $\eta= 5\times10^{-4}$.

\subsection{Post-processing}
\label{sec:pp}
The results of segmentation  were subjected to 3-D connected component analysis followed by slice-wise 2-D connected component analysis and morphological operations such as binary hole-filling inside the ventricular cavity.

\subsection{Cardiac disease diagnosis}
The goal of the automated cardiac disease diagnosis challenge was to classify the cine MRI-scans of the heart into one of the five groups, namely:- (i) DCM, (ii)HCM, (iii) MINF, (iv) ARV and (v) NOR.

\subsubsection{Feature extraction}
From the ground truth segmentations of training data, several cardiac features pertaining to Left Ventricle (LV),  Right Ventricle (RV) and  Myocardium (MYO) were extracted.
The cardiac features were grouped into two categories, namely primary feature \& derived features.
Primary features were calculated directly from the segmentations and DICOM tag pixel spacing and slice-thickness. 
\begin{enumerate}
\item \textbf{Volume} of LV at End Diastole (ED) and End Systole (ES) Phases.
\item \textbf{Volume} of RV at ED and ES Phases
\item \textbf{Mass} and \textbf{volume} of MYO estimated at ED and ES Phases respectively.
\item Myocardial wall \textbf{thickness} at each slice.
\end{enumerate}
Derived features were a combination of primary features:
\begin{enumerate}
\item Ejection Fraction (EF) of LV and RV.
\item Ratio of Primary features: $LV:RV$, $MYO:LV$ at ED and ES phases.
\item Variation profile of Myocardial Wall Thickness (MWT) in Short-Axis (SA) and Long-Axis (LA). \ref{app:D} gives the methodology in arriving at these set of features.
\end{enumerate}

\subsubsection{Two-stage ensemble approach for cardiac disease classification}

\begin{figure*}
\begin{center}
\includegraphics[width=\textwidth]{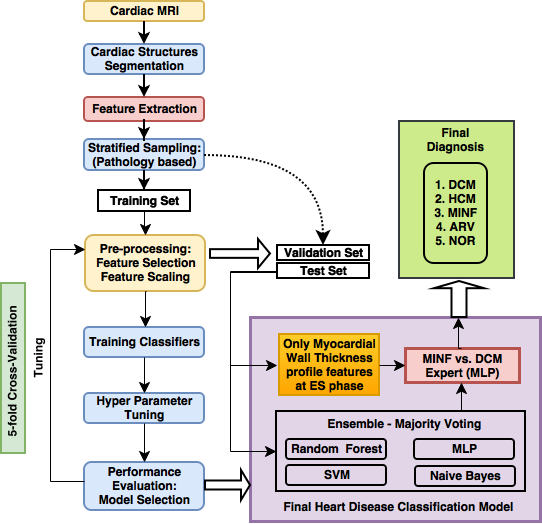}
\end{center}
\caption{Automated Cardiac Disease Diagnosis using Ensemble of Classifiers and Expert Classifier Approach}
\label{fig:disease_diagnosis}
\end{figure*}
The cardiac disease diagnosis was approached as two stage classification using an ensemble system. Figure \ref{fig:disease_diagnosis} illustrates the methodology adopted for cardiac disease classification. Ensemble classification is a process in which multiple classifiers are created and strategically combined to solve a particular a classification problem. Combining multiple-classifiers need not always guarantee better performance than the best individual classifier in the ensemble. The ensemble ensures that overall risk due to poor model selection is minimized. 
The accuracy of the classifiers were estimated based on 5-fold cross-validation scores. Based on the cross-validation scores only the top performing classifiers were selected for combining in the Ensemble-based system. The first stage of the ensemble comprised of four classifiers namely- (i) Support Vector Machine (SVM) with Radial Basis Function kernel, (ii) Multi-layer Perceptron (MLP)  with 2 hidden layers with 100 neurons each, (iii) Gaussian Naive Bayes (GNB) and (iv) Random Forest (RF) with 1000 trees. All the classifiers were independently trained to classify the patient's cine MR scan into five groups by extracting all the features listed in Table \ref{tab:features}. In the first-stage of the Ensemble a voting classifier finalized the disease prediction based on majority vote. 

\begin{table*}
\centering
\caption{The table lists all the features extracted from the predicted segmentation labels. All the 20 features were used for training the classifiers in the first stage of the ensemble. The expert classifier was trained with only a subset of the listed features indicated by $*$. $MWT|SA$:- Set of Myocardial Wall Thickness measures (in mm) per short-axis slice. $Statistic(MWT|SA)|LA$:- Set of statistic (like mean, standard deviation) for all short axis slices when seen along long-axis at a particular cardiac phase}
\label{tab:features}
\resizebox{0.8\textwidth}{!}{%
\begin{tabular}{@{}llll@{}}
\toprule
\textbf{Features} & \textbf{LV} & \textbf{RV} & \textbf{MYO} \\ \midrule
\multicolumn{4}{c}{\textit{\textbf{Cardiac volumetric features}}} \\
\hline
volume at ED & \checkmark & \checkmark &  \\
volume at ES & \checkmark & \checkmark & \checkmark \\
mass at ED &  &  & \checkmark \\
ejection fraction & \checkmark & \checkmark &  \\
volume ratio: $ED[vol (LV) / vol (RV)]$ & \checkmark & \checkmark &  \\
volume ratio: $ES[vol (LV) / vol (RV)]$ & \checkmark & \checkmark &  \\
volume ratio: $ES[vol(MYO) / vol(LV)]$ & \checkmark &  & \checkmark \\
mass to volume ratio: $ED[mass(MYO)/vol(LV)]$ & \checkmark &  & \checkmark \\\hline
\multicolumn{4}{c}{\textit{\textbf{Myocardial wall thickness variation profile}}} \\\hline
$ED[max(mean(MWT|SA)|LA)]$ &  &  & \checkmark \\
$ED[stdev(mean(MWT|SA)|LA)]$ &  &  & \checkmark \\
$ED[mean(stdev(MWT|SA)|LA)]$ &  &  & \checkmark \\
$ED[stdev(stdev(MWT|SA)|LA)]$ &  &  & \checkmark \\
$ES [max(mean(MWT|SA)|LA)] ^{*}$ &  &  & \checkmark \\
$ES[stdev(mean(MWT|SA)|LA)]^{*}$ &  &  & \checkmark \\
$ES[mean(stdev(MWT|SA)|LA)]^{*}$ &  &  & \checkmark \\
$ES[stdev(stdev(MWT|SA)|LA)]^{*}$ &  &  & \checkmark \\ \bottomrule
\end{tabular}%
}
\end{table*}
In some of the cases, the first stage of the ensemble had difficulty in distinguishing between MINF and DCM groups. In-order to eliminate such mis-classifications, a two class $``$ expert" classifier trained only on myocardial wall  thickness variation profile features at ES phase was proposed. The expert classifier re-assessed only those cases for which the first stage's predictions were MINF or DCM. The expert classifier used was MLP with 2 hidden layers with 100 neurons each.
\section{Experimental analysis}
\label{sec:3}
In this section, we experimentally analyze the effectiveness of our proposed network architecture, loss function, data-augmentation scheme, effect of ROI cropping and post-processing. For all the ablation studies we used ACDC training dataset. The metrics of evaluation were Dice score and Hausdorff Distance (HD) in mm (\ref{app:A}). The neural network architectures were designed using TensorFlow (\cite{abadi2016tensorflow}) software. We ran our experiments on a desktop computer with NVIDIA-Titan-X GPU, Intel Core i7-4930K 12-core CPUs @ 3.40GHz and 64GB RAM. 

\subsection{ACDC-2017 dataset}
The training dataset comprising of 100 patient cases ($\approx 1.8k$ 2D images) were split into $70:15:15$ for training, validation and testing. Stratified sampling was done so as to ensure each split comprised of equal number of cases from different cardiac disease groups. Each patient scan had approximately 20 $2D$ images with ground-truth annotations for Left-Ventricle (LV), Right Ventricle (RV) and Myocardium (MYO) at the End Diastole (ED) and End Systole (ES) phases.  
 
\subsubsection{Segmentation network architecture and setting hyper-parameters of the network}
\label{sec:hyp_set}
\begin{figure*}
\centering
\includegraphics[width=0.8\textwidth,keepaspectratio]{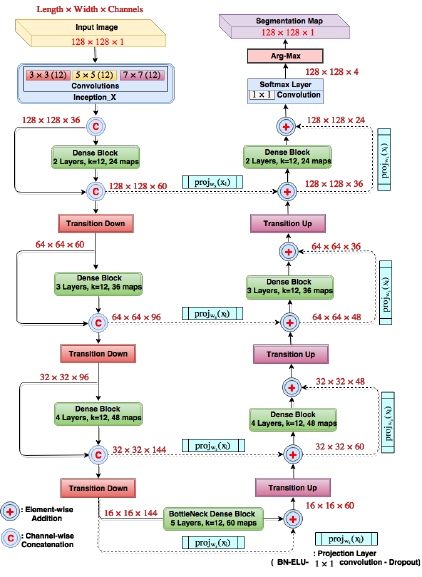}
\caption{The proposed network architecture (DFCN-C) comprises of a contracting path (down-sampling path) and an expanding path (up-sampling path). The arrows indicate network's information flow pathways. The dotted horizontal arrows represent residual skip connections where the feature  maps from the down-sampling path are added in an element-wise manner with the corresponding feature maps in the up-sampling path. In order to match the channel dimensions a linear projection is done using BN-ELU-$1\times1$-convolution-Dropout in the residual connection path. In the down-sampling path, the input to a dense block was concatenated with its output, leading to a linear growth in the number of feature maps. Whereas in the Bottleneck and up-sampling path the features were added element-wise to enable learning a residual function.}
\label{fig:dfcn_c_k_12}
\end{figure*}

Figure \ref{fig:dfcn_c_k_12} describes the proposed network architecture used for segmentation. Based on experimental results, the following network hyper-parameters were fixed:- (i) Number of max-pooling operations were limited to three ($P=3$), (ii) Growth-rate of Dense-Blocks (DBs) was set ($k=12$), (iii) Number of initial feature maps (F) generated by the first convolution layers was ensured to be at-most 3 times the growth-rate ($F \approx 3k$). 

\subsubsection{Training}
\label{sec:train_regime}
The network was trained by minimizing the proposed loss function (Eq. \ref{eq:loss}) using ADAM optimizer (\cite{kingma2014adam}) with a learning rate set to $10^{-3}$. The network weights were initialized using He normal initializer (\cite{he2015delving}) and trained for 200 epochs with data augmentation scheme as described in \cref{sec:aug_details}. The training batch comprised of $16$ ROI cropped 2D MR images of dimension $128\times 128$. After every epoch the model was evaluated on the validation set and the final best model selected for evaluating on test set was ensured to have highest Dice score for MYO class on the validation set.

\subsubsection{Evaluating the effect of growth rate}
Table \ref{tab:growth} shows the DFCN-C performance with varying growth-rate ($k$) parameter. For the same architecture the segmentation performance steadily improved with increasing value of $k$ also the network exhibited potential to work with extremely small number of trainable parameters.

\begin{table*}
\centering
\caption{Evaluation of segmentation results for various growth rates. The values are provided as mean (standard deviation).}
\label{tab:growth}
\resizebox{\textwidth}{!}{%
\begin{tabular}{lllllllll}
\hline
\textbf{} & \multicolumn{8}{c}{\textbf{Growth Rate}} \\ \hline
\textbf{DICE} & \textit{\textbf{k=2}} & \textit{\textbf{k=4}} & \textit{\textbf{k=6}} & \textit{\textbf{k=8}} & \textit{\textbf{k=10}} & \textit{\textbf{k=12}} & \textit{\textbf{k=14}} & \textit{\textbf{k=16}} \\
LV & 0.86 (0.08) & 0.90 (0.06) & 0.88 (0.09) & 0.93 (0.05) & 0.93 (0.04) & 0.93 (0.05) & 0.93 (0.05) & 0.93 (0.05) \\
RV & 0.76 (0.13) & 0.82 (0.14) & 0.85 (0.17) & 0.87 (0.13) & 0.91 (0.04) & 0.91 (0.05) & 0.91 (0.05) & 0.92 (0.05) \\
MYO & 0.78 (0.06) & 0.80 (0.08) & 0.82 (0.09) & 0.88 (0.04) & 0.88 (0.03) & 0.89 (0.03) & 0.90 (0.02) & 0.90 (0.03) \\ \hline
Mean & 0.80 (0.09) & 0.84 (0.09) & 0.85 (0.12) & 0.89 (0.07) & 0.91 (0.04) & 0.91 (0.04) & 0.92 (0.04) & \textbf{0.92 (0.04)} \\ \hline \hline
\textbf{HD} & \textit{\textbf{k=2}} & \textit{\textbf{k=4}} & \textit{\textbf{k=6}} & \textit{\textbf{k=8}} & \textit{\textbf{k=10}} & \textit{\textbf{k=12}} & \textit{\textbf{k=14}} & \textit{\textbf{k=16}} \\
LV & 16.89 (10.51) & 13.04 (9.51) & 17.64 (11.33) & 7.26 (8.97) & 4.96 (5.15) & 5.46 (6.39) & 3.95 (4.21) & 4.16 (4.26) \\
RV & 17.96 (10.86) & 9.41 (4.74) & 7.49 (4.20) & 6.01 (2.65) & 5.60 (2.58) & 5.65 (2.38) & 5.49 (2.82) & 4.79 (2.05) \\
MYO & 14.27 (8.47) & 14.72 (9.57) & 16.71 (8.37) & 5.58 (4.08) & 7.72 (7.80) & 5.18 (4.44) & 4.32 (3.20) & 4.25 (4.12) \\ \hline
Mean & 16.37 (9.95) & 12.39 (7.94) & 13.95 (7.97) & 6.29 (5.24) & 6.09 (5.18) & 5.43 (4.40) & 4.59 (3.41) & \textbf{4.40 (3.48)} \\ \hline
Parameters & 11,452 & 43,036 & 94,756 & 166,612 & 258,604 & 370,732 & 502,996 & 655,396 \\ \hline
\end{tabular}%
}
\end{table*}

\subsubsection{Evaluating the effect of different loss functions}
Table \ref{tab:loss} compares the segmentation performance of six different loss functions. Because of ROI cropping, the heavy class-imbalance was already mitigated and hence the standard cross-entropy loss showed optimal performance in terms of both Dice score and Hausdorff measures. In terms of Hausdorff distance metric alone, the spatially weighted cross-entropy loss showed best performance suggesting heavy weight penalization for contour voxels aided in learning the contours precisely. The performance of standard dice loss was better than the mini-batch weighted dice loss, indicating  weighting was not necessary when using only dice loss. The simple combination of dice and cross-entropy losses showed slight dip in the performance. It was observed that cross-entropy loss optimized for pixel-level accuracy whereas the dice loss helped in improving the segmentation quality/metrics. Weighted dice loss alone caused over segmentation at the boundaries whereas the weighted cross-entropy loss alone led to very sharp contours with minor under-segmentation. So, in order to balance between these trade-offs and combine the advantages of both the losses, we empirical observed that by setting $\gamma = 1$ and $\lambda = 1$ in the proposed loss (Eq. (\ref{eq:loss})) gave optimal performance in terms of faster convergence and better segmentation quality.

\begin{table*}
\centering
\caption{Evaluation of segmentation results for different loss functions. Note:- CE: cross-entropy loss, D: dice loss, sW-CE : cross-entropy loss with weighting scheme based on spatial weight map, mW-D: dice loss with weighting scheme based on mini-batch. The values are provided as mean (standard deviation).}
\label{tab:loss}
\resizebox{\textwidth}{!}{%
\begin{tabular}{@{}llllllll@{}}
\toprule
\textbf{Method} & \multicolumn{2}{l}{\textbf{DICE LV}} & \multicolumn{2}{l}{\textbf{DICE RV}} & \multicolumn{2}{l}{\textbf{DICE MYO}} & \textbf{Mean Dice} \\ \midrule
 & \textbf{ED} & \textbf{ES} & \textbf{ED} & \textbf{ES} & \textbf{ED} & \textbf{ES} &  \\
\textbf{CE} & 0.96 (0.02) & 0.91 (0.07) & 0.94 (0.02) & \textbf{0.88 (0.06)} & 0.87 (0.03) & 0.88 (0.03) & \textbf{0.91 (0.04)} \\
\textbf{sW-CE} & 0.96 (0.02) & \textbf{0.91 (0.06)} & 0.92 (0.09) & 0.85 (0.17) & \textbf{0.89 (0.03)} & 0.89 (0.03) & 0.90 (0.07) \\
\textbf{D} & 0.96 (0.02) & 0.91 (0.09) & 0.95 (0.02) & 0.88 (0.07) & \textbf{0.89 (0.03)} & \textbf{0.90 (0.03)} & \textbf{0.91 (0.04)} \\
\textbf{mW-D} & 0.96 (0.02) & 0.90 (0.08) & \textbf{0.95 (0.01)} & 0.87 (0.06) & 0.87 (0.02) & 0.89 (0.03) & 0.90 (0.04) \\
\textbf{CE+D} & 0.96 (0.02) & 0.90 (0.08) & 0.93 (0.04) & 0.86 (0.11) & 0.88 (0.03) & 0.88 (0.04) & 0.90 (0.05) \\
\textbf{sW-CE + mW-D} & 0.96 (0.02) & 0.90 (0.08) & 0.95 (0.02) & 0.87 (0.08) & \textbf{0.89 (0.03)} & 0.89 (0.03) & \textbf{0.91 (0.04)} \\ \hline \hline
\textbf{} & \textbf{HD LV} & \textbf{} & \textbf{HD RV} & \textbf{} & \textbf{HD MYO} &  & \textbf{Mean HD} \\\midrule
\textbf{} & \textbf{ED} & \textbf{ES} & \textbf{ED} & \textbf{ES} & \textbf{ED} & \textbf{ES} &  \\
\textbf{CE} & \textbf{2.98 (2.93)} & 4.73 (3.78) & 4.81 (1.94) & 6.73 (3.51) & \textbf{3.57 (2.55)} & 7.90 (7.12) & 5.12 (3.64) \\
\textbf{sW-CE} & 3.82 (4.01) & \textbf{4.04 (2.51)} & 5.09 (2.62) & \textbf{6.19 (3.93)} & 4.46 (2.82) & \textbf{4.85 (2.83)} & \textbf{4.74 (3.12)} \\
\textbf{D} & 4.70 (6.58) & 7.82 (9.72) & 4.82 (1.79) & 6.41 (3.26) & 4.59 (4.93) & 6.16 (5.44) & 5.75 (5.29) \\
\textbf{mW-D} & 4.49 (8.35) & 7.45 (9.01) & 10.14 (8.10) & 9.62 (6.65) & 6.47 (7.38) & 9.54 (10.37) & 7.95 (8.31) \\
\textbf{CE+D} & 7.09 (10.24) & 9.47 (10.86) & 4.85 (2.08) & 6.93 (2.88) & 7.67 (8.90) & 11.56 (9.27) & 7.93 (7.37) \\
\textbf{sW-CE + mW-D} & 4.42 (6.39) & 6.51 (6.40) & \textbf{4.20 (1.81)} & 7.10 (2.95) & 4.48 (4.52) & 5.87 (4.35) & 5.43 (4.40) \\ \bottomrule
\end{tabular}%
}
\end{table*}
\subsubsection{Evaluating the effect of ROI cropping and post-processing}
\begin{figure*}
\subfloat[Before LCCA]{\includegraphics[width=0.23\textwidth,height=3cm,]{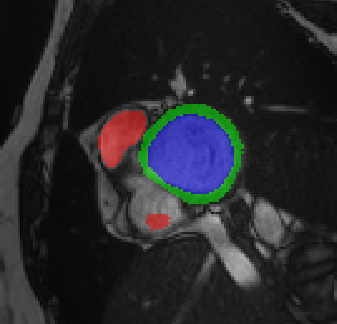}}\hfill
\subfloat[After LCCA]{\includegraphics[width=0.23\textwidth, height=3cm]{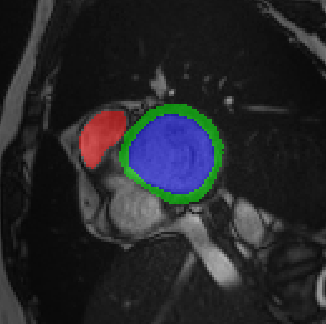}} \hfill
\subfloat[Before MBHF]{\includegraphics[width=0.23\textwidth,height=3cm]{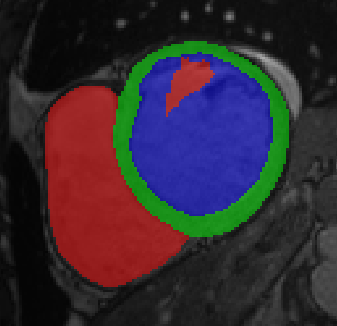}} \hfill
\subfloat[After MBHF]{\includegraphics[width=0.23\textwidth,height=3cm]{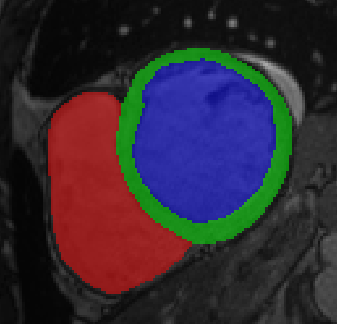}}\\

\caption{The figure shows the sequence of post-processing steps applied to eliminate false-positives and outliers in the predictions. Largest component component analysis (LCCA) retained only the largest common structure and discarded the rest as seen in (a) and (b). Morphological binary hole filling (MBHF) operation eliminated outliers as seen in (c) and (d)}
\label{fig:post_process}
\end{figure*}

For evaluating the effect of not using ROI cropped images, the proposed network was trained on the input images resized to  $256\times256$ (zero-padding or center-cropping was done to ensure this image dimension). 
Table \ref{tab:roi_pp} compares the effect of using ROI-cropping and post-processing on segmentation results. Even though the non-ROI based technique resulted in better dice-score but it had higher Hausdorff distance, this was mainly because of of false positives at basal and apical slices. As shown in Fig. \ref{fig:post_process} post-processing steps aided in removing false positives and outliers. Table \ref{tab:roi_pp} indicates that the performance of ROI vs. non-ROI was comparable after post-processing steps.
However, in the interest of reducing the GPU memory foot-print and time required for training and inference, ROI based method was proposed.
\begin{table*}
\centering
\caption{Evaluation results with and without ROI cropping and post-processing. The values are provided as mean (standard deviation). ROI - Region of Interest, PP - Post-processing.}
\label{tab:roi_pp}
\resizebox{\textwidth}{!}{%
\begin{tabular}{@{}lllllllll@{}}
\toprule
\textbf{ROI} & \textbf{PP} & \multicolumn{2}{l}{\textbf{DICE LV}} & \multicolumn{2}{l}{\textbf{DICE RV}} & \multicolumn{2}{l}{\textbf{DICE MYO}} & \textbf{Mean Dice} \\ \midrule
 &  & \textbf{ED} & \textbf{ES} & \textbf{ED} & \textbf{ES} & \textbf{ED} & \textbf{ES} &  \\
\checkmark & \xmark & 0.96 (0.02) & 0.90 (0.08) & 0.94 (0.04) & 0.84 (0.13) & 0.88 (0.03) & 0.88 (0.03) & 0.90 (0.05) \\
\checkmark & \checkmark & 0.96 (0.02) & 0.90 (0.08) & \textbf{0.95 (0.02)} & \textbf{0.87 (0.08)} & 0.89 (0.03) & 0.89 (0.03) & \textbf{0.91 (0.04)} \\
\xmark & \xmark & 0.96 (0.02) & \textbf{0.91 (0.07)} & 0.93 (0.03) & 0.84 (0.10) & 0.89 (0.02) & 0.90 (0.03) & 0.91 (0.04) \\
\xmark & \checkmark & 0.96 (0.02) & \textbf{0.91 (0.07)} & 0.94 (0.03) & 0.86 (0.08) & \textbf{0.90 (0.02)} & \textbf{0.90 (0.02)} & \textbf{0.91 (0.04)} \\ \hline \hline
 &  & \multicolumn{2}{l}{\textbf{HD LV}} & \multicolumn{2}{l}{\textbf{HD RV}} & \multicolumn{2}{l}{\textbf{HD MYO}} & \textbf{Mean HD} \\ \hline
 &  & \textbf{ED} & \textbf{ES} & \textbf{ED} & \textbf{ES} & \textbf{ED} & \textbf{ES} &  \\
\checkmark & \xmark & 8.04 (11.26) & 15.28 (13.45) & 11.80 (19.17) & 13.34 (11.92) & 11.31 (12.84) & 17.52 (14.10) & 12.88 (13.79) \\
\checkmark & \checkmark & 4.42 (6.39) & 6.51 (6.40) & \textbf{4.20 (1.81)} & 7.10 (2.95) & 4.48 (4.52) & \textbf{5.87 (4.35)} & 5.43 (4.40) \\
\xmark & \xmark & \textbf{2.97 (3.15)} & 9.69 (16.88) & 26.90 (42.44) & 21.49 (35.86) & 22.11 (34.04) & 34.55 (44.34) & 19.62 (29.42) \\
\xmark & \checkmark & \textbf{2.97 (3.15)} & \textbf{3.92 (2.71)} & 4.67 (1.96) & \textbf{7.05 (4.29)} & \textbf{3.72 (2.58)} & 6.09 (6.22) & \textbf{4.74 (3.48)} \\ \bottomrule
\end{tabular}%
}
\end{table*}

\subsubsection{Evaluating the effect of data augmentation scheme}
\label{sec:aug_details}
\begin{figure*}
\centering
\includegraphics[width=\textwidth]{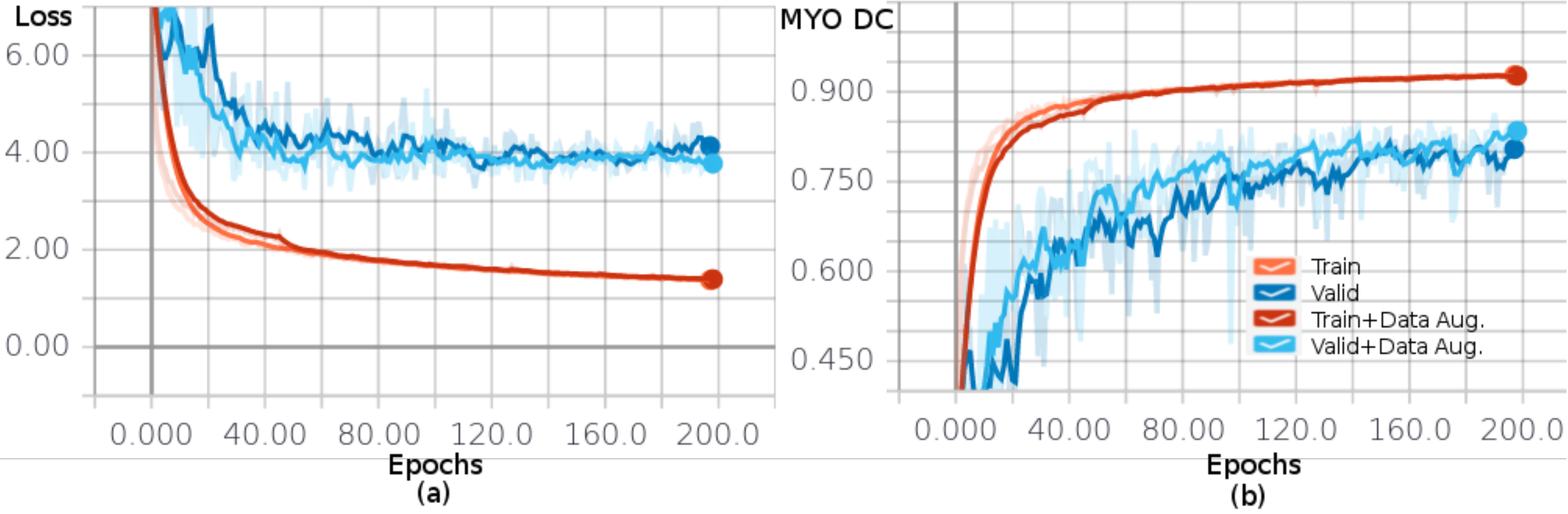}
\caption{Comparison of the learning curves of the DFCN-C with and without data augmentation. (a) Loss curves, (b) Dice score curves for MYO class.}
\label{fig:data_aug}
\end{figure*}

Data augmentation is done to artificially increase the training set and to prevent the network from over-fitting on the training set. For analyzing the effect of data augmentation during the learning process of the proposed network, two separate models were trained with and without data-augmentation. For the model which incorporated data-augmentation scheme the training batch comprised of the mixture of original dataset and on the fly randomly generated augmented data which included: (i) rotation: random angle between $-5$ and $5$ degrees, (ii) translation x-axis: random shift between $-5$ and $5$ mm, (iii) translation y-axis: random shift between $-5$ and $5$ mm, (iv) rescaling: random zoom factor between $0.8$ and $1.2$, (v) adding Gaussian noise with zero mean and 0.01 standard deviation and (vi) elastic deformations using a dense deformation field obtained through a  $2\times2$ grid of control-points and B-spline interpolation.
Figure \ref{fig:data_aug} summarizes the learning curves, both the models validation loss decreased consistently as the training loss went down, indicating less over-fitting on the training data. Closer observation on the validation curves with model trained on data-augmentation revealed minor improvement in the segmentation performance on the validation set which was also corroborated on the held-out test-set.
\subsubsection{Comparing performance and learning curves with other baseline models}
For analyzing and benchmarking the proposed network DFCN-C (F=36, P=3, k=12), we constructed 3 baseline models, namely:- 
(i) Modified U-NET (\cite{ronneberger2015u}) starting with 32 initial feature maps and 3 max-pooling layers (F=32, P=3), (ii) DFCN-A (\cite{jegou2017one}) (F=32, P=3, k=12), (iii) DFCN-B (F=32, P=3, k=12). The architecture of DFCN-B was similar to DFCN-C with only exception being at the initial layer. Its initial layer had only one CNN branch which learnt 32 $3\times 3$ filters. All the models were trained in the same manner. 

\begin{figure*}
\centering
\includegraphics[width=\textwidth]{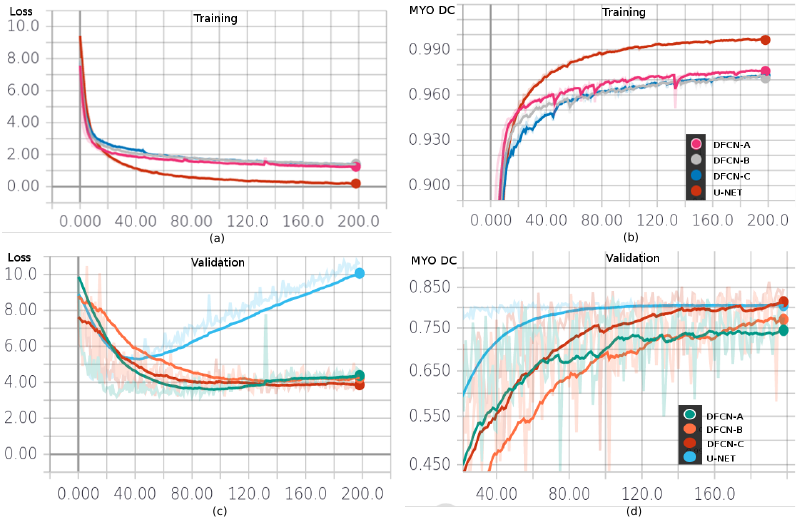}
\caption{Comparison of the learning curves of the DFCN-A, DFCN-B, DFCN-C (proposed) and U-NET. (a) Training loss curves, (b) Dice score curves for MYO class during training, (c) Validation loss curves, (d) Dice score curves for MYO class during validation.}
\label{fig:loss_curves}
\end{figure*}

Figure \ref{fig:loss_curves} summarizes and compares the learning process of DFCN-C with other three baseline models. In U-NET, it was observed that the loss associated with training and validation  to decrease and increase respectively as the training progressed. Such patterns indicate the possibility of the network to over-fit on smaller datasets.   
Moreover, the number of parameters and GPU memory usage were highest for U-NET.
For all the DFCN based architectures, the validation loss consistently decreased as the training loss decreased, hence showed least tendency to over-fit. When comparing amongst the DFCN variants, the validation curves of DFCN-C showed faster convergence and better segmentation scores when compared DFCN-B. Hence corroborating the effectiveness of proposed methodology of multi-scale feature extraction and feature fusion. When compared to DFCN-A, the number of parameters and GPU memory usage were relatively low for both DFCN-B and DFCN-C. Table \ref{tab:arch_compare} compares the results of the proposed DFCN-C architecture against the baseline models.

\begin{table*}
\centering
\caption{Evaluation results for baseline models and DFCN-C. The values are provided as mean (standard deviation). The GPU memory usage shown in the table is when with input images are of dimension $128\times 128$ and the batch-size is $1$.}
\label{tab:arch_compare}
\resizebox{\textwidth}{!}{%
\begin{tabular}{@{}lllllllll@{}}
\toprule
\textbf{Method} & \multicolumn{2}{l}{\textbf{DICE LV}} & \multicolumn{2}{l}{\textbf{DICE RV}} & \multicolumn{2}{l}{\textbf{DICE MYO}} & \textbf{Mean Dice} & \textbf{} \\ \midrule
 & \textbf{ED} & \textbf{ES} & \textbf{ED} & \textbf{ES} & \textbf{ED} & \textbf{ES} &  & \textbf{Params.} \\
\textbf{U-NET} & 0.94 (0.02) & 0.90 (0.06) & 0.90 (0.04) & 0.82 (0.07) & 0.85 (0.05) & 0.86 (0.04) & 0.88 (0.05) & 1,551k \\
\textbf{DFCN-A} & \textbf{0.96 (0.02)} & \textbf{0.91 (0.08)} & 0.94 (0.02) & \textbf{0.88 (0.08)} & \textbf{0.89 (0.03)} & \textbf{0.90 (0.03)} & \textbf{0.91 (0.04)} & 435k \\
\textbf{DFCN-B} & \textbf{0.96 (0.02)} & 0.90 (0.08) & 0.92 (0.05) & 0.84 (0.17) & 0.88 (0.04) & 0.88 (0.05) & 0.90 (0.07) & 360k \\
\textbf{DFCN-C} & \textbf{0.96 (0.02)} & 0.90 (0.08) & \textbf{0.95 (0.02)} & 0.87 (0.08) & \textbf{0.89 (0.03)} & 0.89 (0.03) & \textbf{0.91 (0.04)} & 371k \\ \hline \hline
 & \textbf{HD LV} & \textbf{} & \textbf{HD RV} & \textbf{} & \textbf{HD MYO} &  & \textbf{Mean HD} &  \\\hline
 & \textbf{ED} & \textbf{ES} & \textbf{ED} & \textbf{ES} & \textbf{ED} & \textbf{ES} &  & \textbf{GPU M.} \\
\textbf{U-NET} & 10.99 (12.29) & 11.90 (10.56) & 12.56 (8.39) & 12.67 (9.42) & 9.64 (6.85) & 12.22 (8.08) & 11.66 (9.26) & 3GB \\
\textbf{DFCN-A} & 6.85 (10.45) & \textbf{4.00 (2.73)} & 4.59 (2.09) & \textbf{5.24 (1.67)} & 8.16 (8.95) & \textbf{4.65 (3.17)} & 5.58 (4.84) & 2GB \\
\textbf{DFCN-B} & 8.28 (10.24) & 6.68 (7.37) & 5.19 (2.12) & 7.88 (4.91) & 8.15 (7.51) & 9.18 (7.19) & 7.56 (6.56) & 1GB \\
\textbf{DFCN-C} & \textbf{4.42 (6.39)} & 6.51 (6.40) & \textbf{4.20 (1.81)} & 7.10 (2.95) & \textbf{4.48 (4.52)} & 5.87 (4.35) & \textbf{5.43 (4.40)} & 1GB \\ \bottomrule
\end{tabular}%
}
\end{table*}

\subsubsection{Visualization of initial layer kernels, feature maps and posterior probabilities}
\begin{figure*}
\centering
\includegraphics[width=\textwidth]{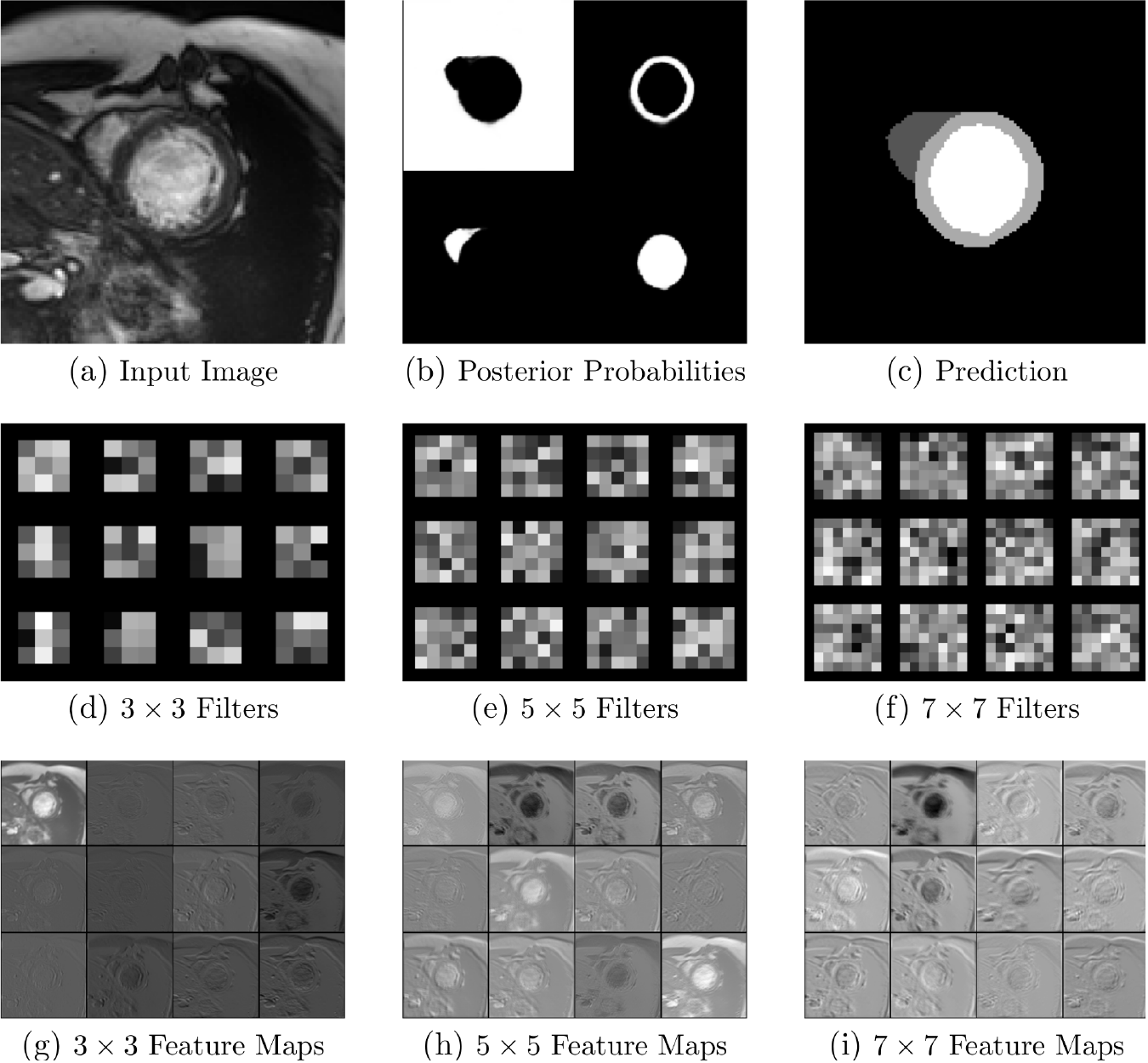}
\caption{The figures compares illustrates the feature maps of trained model. (a) Input image fed to the network, (b) posterior probability maps after soft-max output, (c) The final prediction of labels, (d) - (f) visualization of the initial layers  kernels- $3\times3$, $5\times5$ and $7 \times 7$, (g) - (i) Filter response to the input image (a). }
\label{fig:fmaps_vis}
\end{figure*}

Figure \ref{fig:fmaps_vis} visualizes intermediate feature maps of the trained DFCN-C network on ACDC-2017 dataset. Visualization of initial layer's learned kernels were not intuitively interpretable, but their corresponding feature maps showed distinct pattern for different kernel sizes. In the $3\times 3$ feature maps the edges appeared sharp indicating attention on smaller regions and one of its kernels learnt the identity transformation. Whereas, feature maps of larger kernels had blurred out edges indicating attention over larger region. The posterior probability maps of all the four classes had sharp contours and were not ambiguous. 

\subsubsection{Classifier selection and feature importance study for cardiac disease diagnosis}
\begin{figure*}
\centering
\subfloat[5-Class Feature Importance]{\includegraphics[width=0.5\textwidth, keepaspectratio]{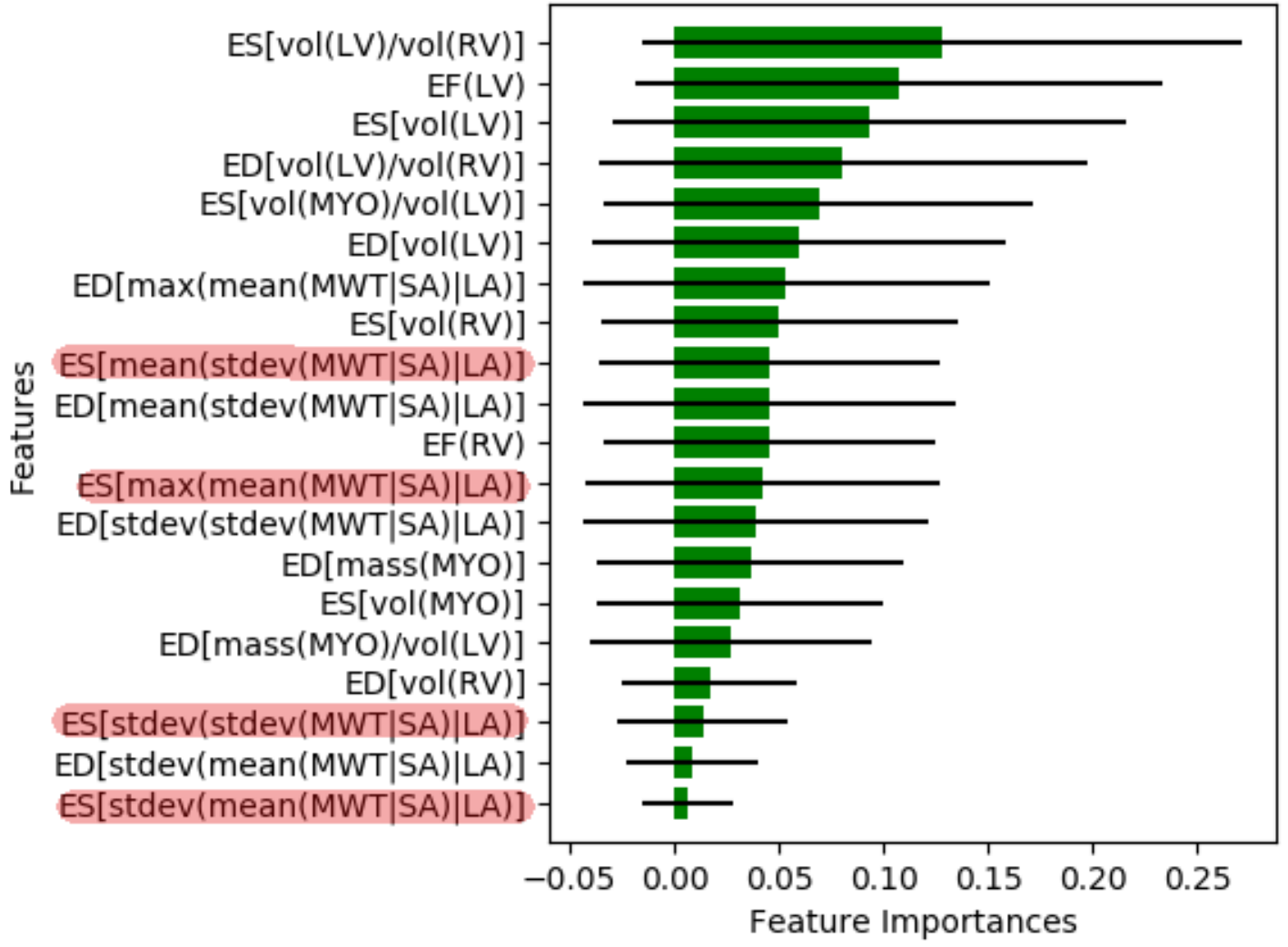}}\hfill
\subfloat[2-Class Feature Importance]{\includegraphics[width=0.5\textwidth, keepaspectratio]{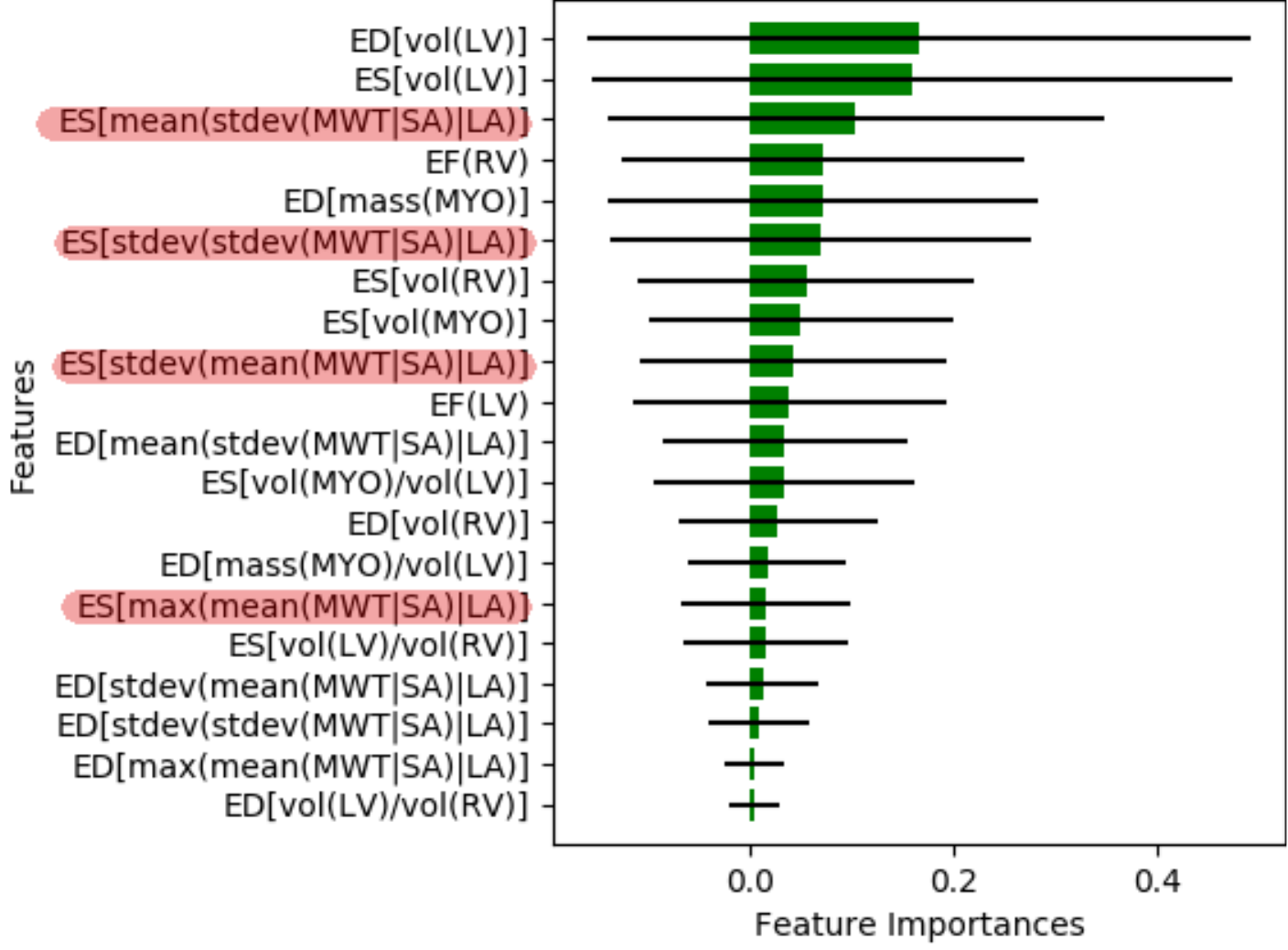}}\\
\caption{Feature importance ranking by Random Forest classifier for two different classification tasks. The green bars are the feature importances of the forest, along with their inter-trees variability (standard deviation). The features highlighted in red color indicate hand-crafted myocardial wall thickness features at ES phase. (a) shows the feature importance for 5-class task, it can be seen that highlighted features have been given low importance, (b) shows the feature importance for the 2-class task, clearly it can be seen that the highlighted features have been ranked higher.}
\label{fig:fimp}
\end{figure*}
In order to validate the hypothesis that myocardial features alone were sufficient for distinguishing between MINF and DCM, feature importance study was done using Random Forest classifier trained to classify between MINF and DCM cases using all the features listed in Table \ref{tab:features}. The Fig. \ref{fig:fimp}((a) \& (b)) compares the feature importance ranking by Random forest when trained to classify all the five groups vs. only DCM and MINF respectively. The expert classifier was trained on myocardial features at End Systole phase only. So, Table \ref{tab:cv} lists the classifiers experimented for the First-stage and the Expert stage classification tasks. Only classifiers with accuracy scores above $95\%$ were selected for the ensemble.

\begin{table}
\centering
\caption{The table lists the classifiers evaluated and their corresponding five-fold cross-validation accuracy scores. The classifiers were evaluated for both first stage of the ensemble (5-class) and expert discrimination (2-class). The values are provided as mean (standard deviation). LR- Logistic Regression, RF- Random Forest with 1000 trees, GNB- Gaussian Naive Bayes, XGB- Extreme Gradient Boosting with 1000 trees, SVM- Support Vector Machine with Radial Basis kernel, MLP- Multi-layer perceptron with 2-hidden layers with 100 neurons each, K-NN- 5-Nearest Neighbors, Vote- Voting classifier based on SVM, MLP, NB \& RF.}
\label{tab:cv}
\resizebox{0.45\textwidth}{!}{%
\begin{tabular}{lll}
\hline
\textbf{Classifier} & \textbf{5-class} & \textbf{2-class} \\ \hline
LR & 0.94 (0.06) & 0.82 (0.06) \\
RF & 0.96 (0.02 & 0.85 (0.09) \\
GNB & 0.96 (0.04) & 0.82 (0.06) \\
XGB & 0.93 (0.04) & 0.88 (0.11) \\
SVM & 0.95 (0.04) & 0.85 (0.09) \\
MLP & \textbf{0.97 (0.02)} & \textbf{0.97 (0.05)} \\
K-NN & 0.91 (0.04) & 0.85 (0.09) \\
Vote & 0.97 (0.04) & 0.93 (0.06) \\ \hline
\end{tabular}
}
\end{table}
\subsection{LV-2011 dataset}
The training dataset comprised of 100 patients ($\approx 29k$ 2D images) and it was randomly split into $70:20:10$ for training, validation and testing.
\subsubsection{Network architecture}
The network architecture used for training on LV-2011 dataset was as per \cref{sec:hyp_set}). The network in Fig. \ref{fig:dfcn_c_k_12} was modified to reflect the number of classes (n=2) in the final classification layer.
\subsubsection{Training}
The model was trained for 50 epochs as described in \cref{sec:train_regime}. Additionally the data-augmentation scheme included random vertical and horizontal flips. 

\section {Results}
\label{sec:4}
\subsection{Performance evaluation on ACDC challenge}
In order to gauge performances on the held out test set by the challenge organizers, clinical metrics such as the average ejection fraction (EF) error, the average  left ventricle (LV) and right ventricle (RV) systolic and diastolic volume errors, and the average myocardium (MYO) mass error were used. For the geometrical metrics, the Dice and the Hausdorff distances for all 3 regions at the ED and ES phases were evaluated. For the cardiac disease diagnosis the metrics used was accuracy.

\subsubsection{Segmentation results}
\begin{table*}
\centering
\caption{Comparisons with different approaches on cardiac segmentation. The evaluations of Left ventricle, Right Ventricle and Myocardium are listed in top, middle and bottom, respectively. Our proposed method had an overall ranking of 2. The values provided are mean (standard deviation). DC- Dice score, HD- Hausdorff distance, cor- correlation}
\label{tab:segm_acdc}
\resizebox{\textwidth}{!}{%
\begin{tabular}{@{}llllllllllll@{}}
\toprule
\textbf{\begin{tabular}[c]{@{}l@{}}Rank\end{tabular}} & \textbf{Method} & \textbf{\begin{tabular}[c]{@{}l@{}}DC\\ ED\end{tabular}} & \textbf{\begin{tabular}[c]{@{}l@{}}DC\\ ES\end{tabular}} & \textbf{\begin{tabular}[c]{@{}l@{}}HD\\ ED\end{tabular}} & \textbf{\begin{tabular}[c]{@{}l@{}}HD\\ ES\end{tabular}} & \textbf{EF cor.} & \textbf{\begin{tabular}[c]{@{}l@{}}EF \\ bias\end{tabular}} & \textbf{EF std.} & \textbf{\begin{tabular}[c]{@{}l@{}}Vol. ED\\ corr.\end{tabular}} & \textbf{\begin{tabular}[c]{@{}l@{}}Vol. ED\\ bias\end{tabular}} & \textbf{\begin{tabular}[c]{@{}l@{}}Vol. ED\\ std.\end{tabular}} \\ \midrule 
1 & \cite{isensee2017automatic} & 0.968 & 0.931 & 7.384 & 6.905 & 0.991 & 0.178 & 3.058 & 0.997 & 2.668 & 5.726 \\
\textbf{2} & \textbf{Ours} & 0.964 & 0.917 & 8.129 & 8.968 & 0.989 & -0.548 & 3.422 & \textbf{0.997} & 0.576 & \textbf{5.501} \\
3 & Yeonggul Jang & 0.959 & 0.921 & 7.737 & 7.116 & 0.989 & -0.330 & 3.281 & 0.993 & -0.440 & 8.701 \\
4 & \cite{baumgartner2017exploration} & 0.963 & 0.911 & 6.526 & 9.170 & 0.988 & 0.568 & 3.398 & 0.995 & 1.436 & 7.610 \\\hline \hline

1 & \cite{isensee2017automatic} & 0.946 & 0.899 & 10.123 & 12.146 & 0.901 & -2.724 & 6.203 & 0.988 & 4.404 & 10.823 \\
2 & \cite{zotti2017gridnet} & 0.941 & 0.882 & 10.318 & 14.053 & 0.872 & -2.228 & 6.847 & 0.991 & -3.722 & 9.255 \\
3 & \textbf{Ours} & 0.935 & 0.879 & 13.994 & 13.930 & 0.858 & -2.246 & 6.953 & 0.982 & -2.896 & 12.650 \\
4 & \cite{baumgartner2017exploration} & 0.932 & 0.883 & 12.670 & 14.691 & 0.851 & 1.218 & 7.314 & 0.977 & -2.290 & 15.153 \\ \hline \hline

1 & \cite{isensee2017automatic} & 0.902 & 0.919 & 8.720 & 8.672 & 0.985 & -3.842 & 9.153 & 0.989 & -4.834 & 7.576 \\
\textbf{2} & \textbf{Ours} & 0.889 & 0.898 & 9.841 & 12.582 & 0.979 & -2.572 & 11.037 & \textbf{0.990} & -2.873 & \textbf{7.463} \\
3 & \cite{baumgartner2017exploration} & 0.892 & 0.901 & 8.703 & 10.637 & 0.983 & -9.602 & 9.932 & 0.982 & -6.861 & 9.818 \\
4 & \cite{patravali20172d} & 0.882 & 0.897 & 9.757 & 11.256 & 0.986 & -4.464 & 9.067 & 0.989 & -11.586 & 8.093 \\\bottomrule
\end{tabular}%
}
\end{table*}
\begin{figure*}
\centering
\includegraphics[width=0.6\textwidth]{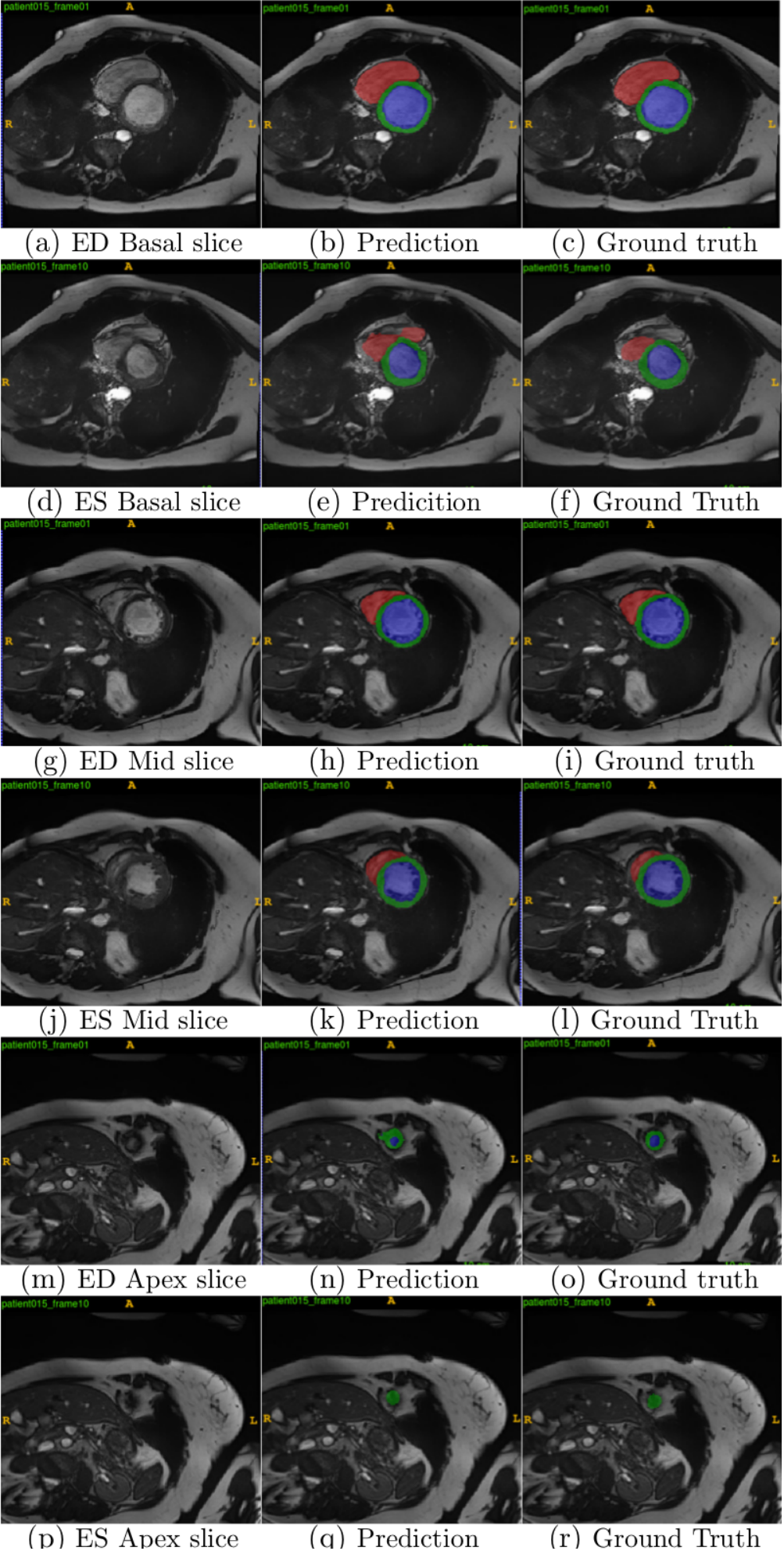}
\caption{Segmentation results at both ED\& ES phases of cardiac cycles on a subset of ACDC training set reserved for testing. The columns from left to right indicate: the input images, segmentations generated by the model and their associated ground-truths. The rows from top to bottom indicate: short axis slices of the heart at basal, mid and apex. In all figures the colors red, green and blue indicate RV, MYO and LV respectively. The model did erroneous segmentation for RV in the basal slice (e) and the Myocardium was over-segmented in the apical slice (n).}
\label{fig:ED_ES}
\end{figure*}
Table \ref{tab:segm_acdc} summarizes our segmentation results along with other top three performing methods. Our method was ranked 2nd for LV and MYO and 3rd for RV. Our method relied on automated localization of the LV center and cropping a patch of size $128\times 128$ around it. So, in cases of abnormally large RV, the model under-segmented RV region when its area extended beyond the patch size. The aforementioned reasons \& irregular shape of RV when compared to LV lead to a dip in dice score \& higher Hausdorff distance for RV  as seen in MICCAI 2017 leader-board (Table \ref{tab:segm_acdc})). Figure \ref{fig:ED_ES} shows the results of segmentation produced by our method at ED and ES phase of the cardiac cycle on held-out training dataset reserved for testing. Our network produced accurate results on most of the slices, but sometimes gave  erroneous segmentation at basal and apical slices.
All the top performing methods were based on CNNs and the top performing method (\cite{isensee2017automatic}) employed ensembling and label-fusion strategy to combine the results of multiple models based on 2D and 3D U-NET inspired architectures.

\subsubsection{Automated disease diagnosis results}
  Table \ref{tab:diag_acdc} summarizes our automated cardiac disease diagnosis results along with other methods. Our 2-stage disease classification approach along with hand-crafted features for myocardial wall thickness helped in surpassing other methods. Other methods mostly relied on cardiac volumetric features and single stage classification model for automated cardiac diagnosis.
  
\begin{table}
\centering
\caption{Comparisons with different approaches on automated cardiac disease diagnosis. Our accuracy score reported in the table is from post-MICCAI leader board.}
\label{tab:diag_acdc}
\resizebox{0.4\textwidth}{!}{%
\begin{tabular}{lll}
\hline
\textbf{Rank} & \textbf{Method} & \textbf{Accuracy} \\ \hline
\textbf{1} & \textbf{Ours} & \textbf{1} \\
2 & \cite{isensee2017automatic} & 0.92 \\
3 & \cite{wolterink2017automatic} & 0.86 \\
4 & Lisa Koch & 0.78 \\ \hline
\end{tabular}%
}
\end{table}
\subsection{Performance evaluation on LV-2011 Challenge}
The challenge organizers evaluated our segmentation results on those images of final validation set for which reference consensus contours CS* \cite{suinesiaputra2014collaborative} were available. The organizers categorized individual images of the final validation set into apex, mid and basal slices. The reported metrics were Jaccard index, Dice score, Sensitivity, Specificity, Accuracy, Positive Predictive Value and Negative Predictive Value (refer \ref{app:A}).

\subsubsection{Segmentation results}
\begin{table*}
\centering
\caption{Comparison of segmentation performance with other published approaches on LV2011 validation set using the consensus contours. AU (\cite{li2010line}), AO (\cite{fahmy2011myocardial}), SCR (\cite{jolly2011automatic}), DS, and INR (\cite{margeta2011layered}) values were taken from Table 2 of (\cite{suinesiaputra2014collaborative} ). FCN values are taken from Table 3 of ( \cite{tran2016fully} ) and CNR regression was taken from Table 3 of \cite{tan2017convolutional}. Values are provided as mean (standard deviation) and in descending order by Jaccard index. FA- Fully Automated}
\label{tab:LV2011}
\resizebox{0.8\textwidth}{!}{%
\begin{tabular}{lllllll}
\hline
\textbf{Method} & \textbf{FA} & \textbf{Jaccard} & \textbf{Sensitivity} & \textbf{Specificity} & \textbf{PPV} & \textbf{NPV} \\ \hline
AU & \xmark & 0.84 (0.17) & 0.89 (0.13) & 0.96 (0.06) & 0.91 (0.13) & 0.95 (0.06) \\
CNR & \xmark & 0.77 (0.11) & 0.88 (0.09) & 0.95 (0.04) & 0.86 (0.11) & 0.96 (0.02) \\
FCN & \checkmark & 0.74 (0.13) & 0.83 (0.12) & 0.96 (0.03) & 0.86 (0.10) & 0.95 (0.03) \\
\textbf{Ours} & \checkmark & 0.74 (0.15) & 0.84 (0.16) & 0.96 (0.03) & 0.87 (0.10) & 0.95 (0.03) \\
AO & \xmark & 0.74 (0.16) & 0.88 (0.15) & 0.91 (0.06) & 0.82 (0.12) & 0.94 (0.06) \\
SCR & \checkmark & 0.69 (0.23) & 0.74 (0.23) & 0.96 (0.05) & 0.87 (0.16) & 0.89 (0.09) \\
DS & \xmark & 0.64 (0.18) & 0.80 (0.17) & 0.86 (0.08) & 0.74 (0.15) & 0.90 (0.08) \\
INR & \checkmark & 0.43 (0.10) & 0.89 (0.17) & 0.56 (0.15) & 0.50 (0.10) & 0.93 (0.09) \\ \hline
\end{tabular}%
}
\end{table*}
\begin{figure*}
\centering
\includegraphics[width=\textwidth, keepaspectratio]{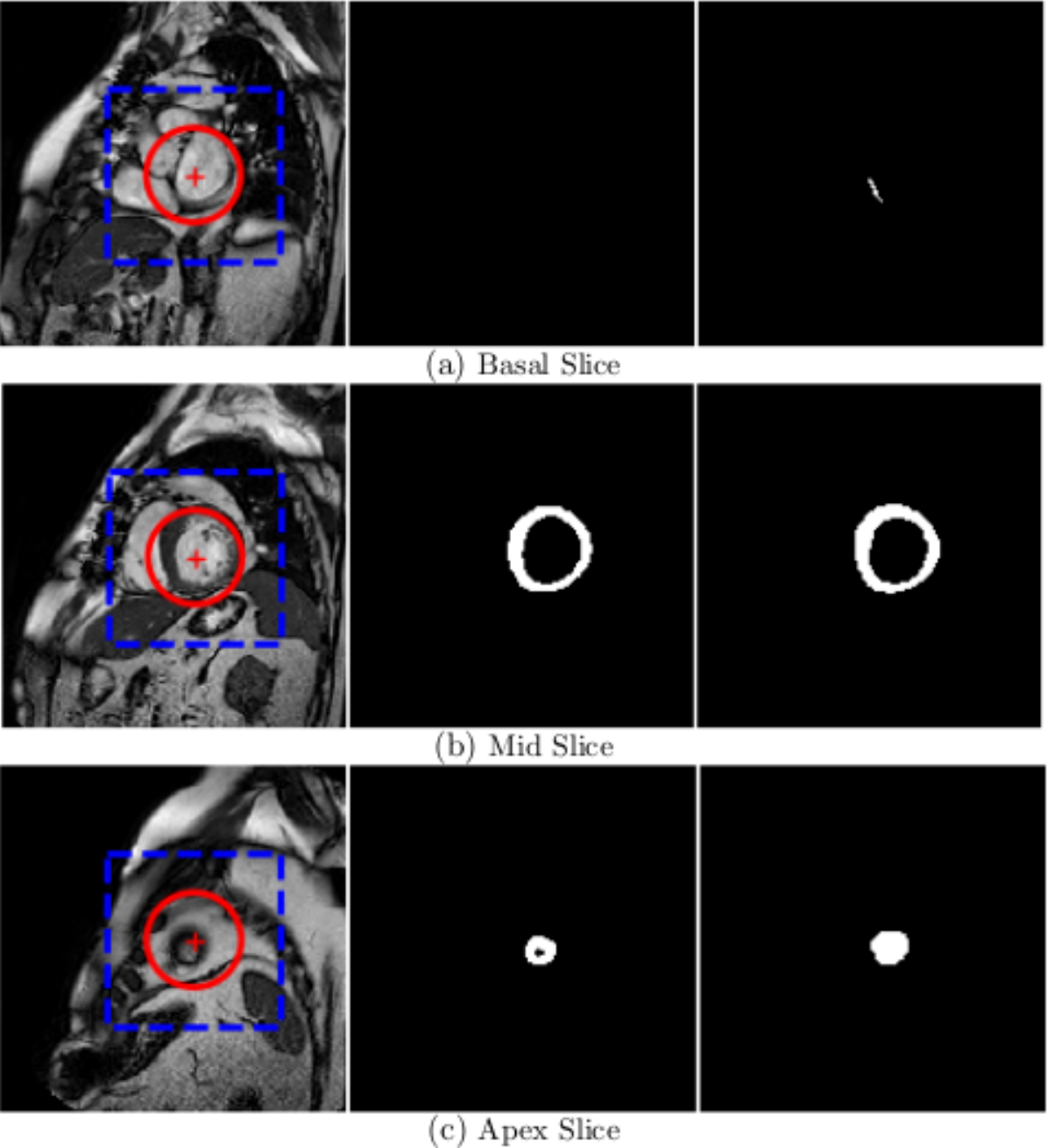}
\caption{From Right to Left: Input, Prediction and Ground Truth for basal, mid and apical slices. The segmentation results are for a subset of LV-2011 training set reserved for testing. The $\oplus$ indicates the ROI center detected by Fourier-Circular Hough Transform approach. The red circle represents the best fitting circle across all slices (Myocardium contours of the mid-slices are more circular in shape and hence the best fitting Hough circle radii mostly encompass them). The blue bounding box indicates the $128\times 128$ patch cropped around ROI center for feeding the segmentation network. Some of the basal slices of the training data had ground truths with partial Myocardium. Even though we trained our model with such slices, our model showed the ability to generalize well.}
\label{fig:lv2011sample}
\end{figure*}

Table \ref{tab:LV2011} compares the results between our proposed approach and other published results using LV-2011 dataset. Our Jaccard index (mean $\pm$ standard deviation) was $0.68 \pm 0.16$, $0.78 \pm 0.13$, $0.74 \pm 0.18$ for the apex, mid and base slices respectively. The errors were mostly concentrated in apical slices. For most of the evaluation measures including Jaccard index, our approach was on par with other fully automated published methods. The AU method used manual guide-point modeling and required human expert approval for all slices and frames. The CNR (\cite{tan2017convolutional}) method used manual input for identifying basal and apical slices and used convolutional regression approach to trace Endocardium and Epicardium contours from the estimated LV center and their network had about $3$ million parameters. When compared to FCN (\cite{tran2016fully}) our network performance was on par with it. But, in contrast to our network architecture and training regime, FCN had about $11$ million parameters and used $95\%$ of the dataset for training. Figure \ref{fig:lv2011sample} illustrates Myocardium prediction at three different slice levels on a subset of training set reserved for testing.
\subsection{Model generalization across different data distributions and effect of sparse annotations}
Table \ref{tab:acdc_only} compares the Myocardium segmentation performance on LV-2011 final validation set between models trained on LV-2011 and ACDC dataset. Despite the ACDC-2017 model being trained to segment LV, RV and MYO at only ED and ES frames with just one-sixth of the training images used to train LV-2011 model, it was able to generalize segmentation for all cardiac frames. The results corroborates the model's effectiveness with sparse annotations and its generalization across datasets arising from completely different data distribution.

\begin{table}
\centering
\caption{Comparison of segmentation performance across different data distributions and effect of sparse annotation. The table compares the myocardium segmentation performance of model trained on only ACDC dataset against model trained on LV-2011 dataset. Both the models were evaluated on LV-2011 final validation set. The values indicate mean (standard deviation)}
\label{tab:acdc_only}
\resizebox{\textwidth}{!}{%
\begin{tabular}{@{}lllllllll@{}}
\toprule
\textbf{Dataset} & \textbf{Training  images} & \textbf{Jaccard} & \textbf{Dice} & \textbf{Accuracy} & \textbf{Sensitivity} & \textbf{Specificity} & \textbf{PPV} & \textbf{NPV} \\ \midrule
LV-2011 & 20,360 & 0.74 (0.15) & 0.84 (0.14) & 0.93 (0.04) & 0.84 (0.16) & 0.96 (0.03) & 0.87 (0.10) & 0.95 (0.03) \\
ACDC-2017 & 1,352 & 0.71 (0.13) & 0.82 (0.11) & 0.92 (0.04) & 0.81 (0.15) & 0.91 (0.06) & 0.82 (0.12) & 0.94 (0.06) \\ \bottomrule
\end{tabular}%
}
\end{table}

\section{Discussion and conclusion}
In this paper, we demonstrated the utility and efficacy of fully convolutional neural network architecture based on Residual DenseNets with Inception architecture in the initial layer for cardiac MR segmentation. Rather than using  ensembles or cascade of networks for segmentation, we showed that a single efficient network trained end-to-end has the potential to learn to segment the Left \& Right Ventricles and Myocardium for all the short-axis slices in all cardiac phases. Our approach achieved near state-of-the-art segmentation results on two benchmark cardiac MR datasets which exhibited variability seen in cardiac anatomical \& functional characteristics across multiple clinical institutions, scanners, populations and heart pathology. Comprehensive evaluations based on multiple metrics revealed that our entire pipeline starting from classical computer vision based techniques for ROI cropping to CNN based segmentation was robust and hence yielded higher segmentation performance. For a $4D$ volume of input dimension $256\times 256 \times 10 \times 30$ it took about 3 seconds for ROI detection and 7 seconds for network inference.

When compared to most of the other published CNN based approaches (\cite{lieman2017fastventricle}, \cite{tran2016fully}, \cite{tan2017convolutional}) for cardiac MR short-axis segmentation our model required least number of trainable parameters ($0.4$ million, an order of 10 fold reduction when compared to standard U-NET based architectures). Our ablation studies indicated that FCN / U-NET based architectures showed higher tendencies to over-fit on small datasets and in absence of proper training strategy like data-augmentation the networks failed to generalize well. Our network's connectivity pattern ensured optimal performance with least model capacity and showed better generalization on small datasets even without employing data-augmentation. By incorporating residual type long skip and short-cut connections in the up-sampling path we overcame the memory explosion seen in FCN based on DenseNets. For multi-scale processing of images, we introduced the idea of incorporating inception style parallelism in the segmentation network. Though we limited the inception structure to only first layer, this could be extended to deeper layers in a computationally efficient manner. The proposed weighting scheme in the dual objective function combined the benefits of both vanilla cross-entropy loss and dice loss.

The main limitation of our model lies in its inability to segment cardiac structures in extremely difficult slices of the heart such as apex and basal regions. Segmentation errors on apical slices have very minor impact on the overall volume computation and hence for cardiac disease diagnosis precise segmentation of all the slices was not always necessary. The demonstrated potential of the our 2D-DFCN model can be scaled to 3-D for volumetric segmentation tasks. Lastly, we present a fully automated  framework which does both cardiac segmentation and precise disease diagnosis, which has potential for usage in clinical applications.


\appendix
\section{Evaluation Metrics}
\label{app:A}
A  brief overview of the main metrics reported in the literature used for comparative purposes is listed in this section.
Let P and G be the set of voxels enclosed by the predicted and ground truth contours delineating the object class in a medical volume respectively. The following evaluation metrics were used to assess the quality of automated segmentation methods using the ground truth as reference:  
\begin{enumerate}
\item \textbf{Dice overlap coefficient} is a metric used for assessing the quality of segmentation maps. It basically measures how similar predicted label maps are with respect to ground truth. The Dice score varies from zero to one (in-case of perfect overlap).
\begin{align}
\label{eq:dice1}
DICE &= \frac{2TP}{2TP + FP + FN}\\
DICE &=\frac{2|P\cap G|}{|P| +|G|}
\end{align}
where $|.|$ denotes the cardinality of the set, $TP$ \& $FP$ indicate True and False Positives, $TN$ \& $FN$:- True and False Negatives. 

The dice coefficient between two binary volumes can be written as:
\begin{align}
\label{eq:dice}
DICE &=\frac{2\sum_{i}^{N}p_ig_i}{\sum_i^Np_i^2+\sum_i^Ng_i^2}
\end{align}
where the sums run over the $N$ voxels, of the predicted binary segmentation volume $p_i \in P$ and the ground truth binary volume $g_i \in G$.

\item \textbf{Jaccard index} (also know as Intersection over Union) is a overlap measure used for comparing the similarity and diversity between two sets. The Jaccard index varies from zero to one (representing perfect correspondence).
\begin{align}
\label{eq:jac}
JACCARD &=\frac{|P\cap G|}{|P \cup G|}
\end{align}

\item Sensitivity (TPR), Specificity (SPC), Positive Predictive Value (PPV) and Negative Predictive Value (NPV) are defined as:
\begin{align}
TPR &= \frac{TP}{(TP+FN)}\\
SPC &= \frac{TN}{(TN+FP)}\\  	
PPV &= \frac{TP}{(TP+FP)}\\  	
NPV &= \frac{TN}{(TN+FN)}
\end{align}

\item \textbf{Hausdorff distance} is a symmetric measure of distance between two contours and is defined as:
\begin{align}
H(P,G) &= \max (h(P,G), h(G,P))\\
h(P,G) &= \max_{p_i \in P}\min_{g_i \in G}||p_i - g_i||
\end{align}
A high Hausdorff value implies that the two contours do not closely match. The Hausdorff distance is computed in millimeter with spatial resolution obtained from the DICOM tag Pixel Spacing.
\end{enumerate}

\section{Region of Interest Detection}
\label{app:B}

\subsection{Fourier Analysis}
The discrete Fourier transform Y of an N-D array X is defined as
\begin{equation}
Y_{K_{1}, K_{2}, \ldots, K_{N}} = \sum_{n_{1}=0}^{m_{1}-1}\omega_{m_{1}}^{K_{1}n_{1}}
								\sum_{n_{2}=0}^{m_{2}-1}\omega_{m_{2}}^{K_{2}n_{2}}\ldots
                                \sum_{n_{N}=0}^{m_{N}-1}\omega_{m_{N}}^{K_{N}n_{N}} X_{n_{1},n_{2}, \ldots,n_{N}}
\label{eq:ndft}
\end{equation}
Each dimension has length $m_{k}$ for $k=1, \,2, \, \cdots , \, N$, and $\omega_{m_{k}}=e^{\frac{âˆ’2 \pi i}{m_{k}}}$ are complex roots of unity where $i$ is the imaginary unit. It can be seen that the N-D Fourier transform Eq.(\ref{eq:ndft}) of an N-D array is equivalent to computing the 1-D transform along each dimension of the N-D array. 

The short-axis cardiac MR images of a slice were taken across entire cardiac cycle and these sequence of images can be treated as 2-dimensional signal varying over time (2D+T signal - $Height\times Width \times Time$). The structures pertaining to heart like the Myocardium and ventricles show significant changes due to heart beat motion. Hence, by taking 3-D Fourier Transform along time axis and analyzing the fundamental Harmonic (also called H1 component, where H stands for Hilbert Space) it was possible to determine pixel regions corresponding to ROI which show strongest response to cardiac frequency. 

\begin{figure*}
\centering
\includegraphics[width=0.5\textwidth]{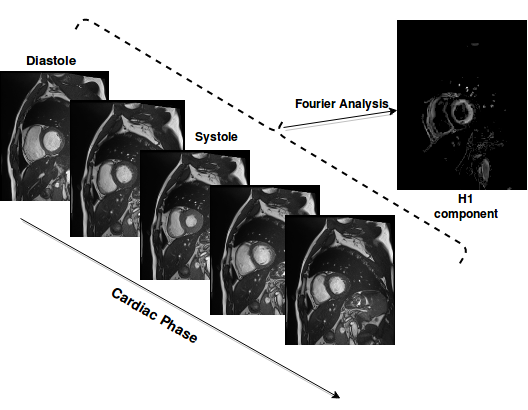}
\caption{The figure shows an example of H1 component image got from a series of MR images of a cardiac short axis slice. It can be seen that most of the chest cavity excluding heart and some adjacent structures have disappeared. If we consider a pixel just outside the LV blood pool at the end diastole, it goes from being bright when it was inside the blood pool to dark when it was in Myocardium at End Systole, because the region containing blood was contracting inwards. The pixel gets brighter again as the heart approaches end diastole. The said pixel's intensity variation will resemble a waveform having frequency same as the heartbeat, hence the H1 component captures those structures of the heart which were responsible for heart beat.}
\label{fig:Fourier_Analysis}
\end{figure*}

The first harmonic of the 3-D FFT was transformed back into original signal's domain (spatial) using 2-D inverse FFT. The result of the previous transformation lead to Complex valued signal. Since, the original signal was Real, the phase component was ignored and only the magnitude of the H1 component (see Figure \ref{fig:Fourier_Analysis}) was retained. The H1 components were estimated for all the slices of the heart starting from base to apex and stacked to form a 3-D volume. The noise present throughout this whole volume was reduced by discarding pixel values which were less than 1\% of the maximum pixel intensity in the whole volume. 

\subsection{Circular Hough Transform}
\begin{figure*}
\centering
\includegraphics[width=1\textwidth, keepaspectratio]{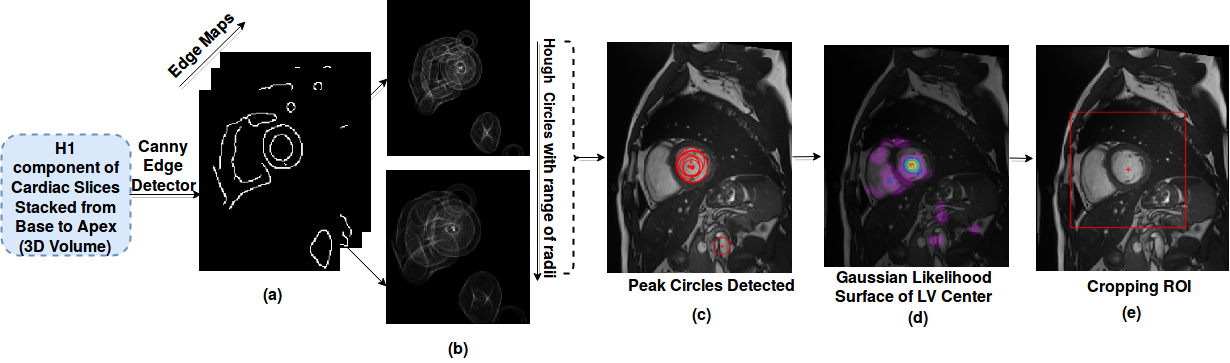}
\caption{Gaussian Kernel-based Circular Hough Transform approach was used for left ventricle (LV) localization. The steps involved in ROI detection were (a) Canny edge detection was done on each short axis slice's H1 component image, (b) For each of the edge maps the Hough circles for a range of radii were found, (c) For each of the edge maps, only $P$ highest scoring Hough circles were retained, where $P$ was a hyper-parameter, (d) For each of the retained circles, votes were cast using an Gaussian kernel that models the uncertainty associated with the circle's center. This approach makes the transform more robust to the detection of spurious circles (in the figure LV center's likelihood surface is overlayed on a slice, the red and purple regions indicates high and low likelihood of LV center respectively), (e) The maximum across LV likelihood surface was selected as the center of the ROI and a square patch of fixed size ($128 \times 128 $) was cropped.}
\label{fig:hough_transform}
\end{figure*}

The LV Myocardium wall resembles circular ring and this contracts and expands during the cardiac cycle. The pixel regions whose intensity varied because of this movement were captured by the H1 component (seen as bright regions in the image). On applying Canny Edge Detection on these H1 component images, two concentric circles were seen which approximate the myocardial wall boundaries at End Diastole and End Systole phases. Henceforth, the localization of the left ventricle center was done using Gaussian Kernel-based Circular Hough transform approach (See Figure \ref{fig:hough_transform}).

\section{Overview of CNN connectivity pattern variants}

\label{app:C}
\subsection{Overview of DenseNets} \label{densenet_overview}
DenseNets are built from dense blocks and pooling operations, where each dense block (DB) is an iterative concatenation of previous feature maps whose sizes match. A layer in dense block is composition of Batch Normalization (BN) (\cite{ioffe2015batch}), non-linearity (activation function), convolution and dropout (\cite{srivastava2014dropout}). The output dimension of each layer has k feature maps where k, is referred to growth rate parameter, is typically set to a small value (e.g. k=8). Thus, the number of feature maps in DenseNets grows linearly with the depth. For each layer in a DenseNet, the feature-maps of all preceding layers of matching spatial resolution are used as inputs, and its own feature-maps are passed onto subsequent layers.
The output of the $l^{th}$ layer is defined as:
\begin{equation}
\label{eq:dense}
x_{l}=H_{l}([x_{l-1},x_{l-2}, \cdots, x_0])
\end{equation}
where $x_{l}$ represents the feature maps at the $l^{th}$ layer and $[\cdots]$ represents the concatenation operation. In our case, H is the layer comprising of Batch Normalization (BN), followed by Exponential Linear Unit (ELU) (\cite{clevert2015fast}), a convolution and dropout rate of $0.2$. A Transition Down (TD) layer is introduced for reducing spatial dimension of feature maps which is accomplished by using a $1\times1$ convolution (depth preserving) followed by a $2\times 2$ max-pooling operation.
This kind of connectivity pattern has the following advantages: \begin{itemize}
\item It ensures that the error signal can be easily back-propagated to earlier layers more directly so this kind of implicit deep super vision, as earlier layers can get more direct supervision from the final classification layer.
\item Higher parameter and computation efficiency is achieved than a normal ConvNet. In a normal ConvNet the number of parameters is proportional to square of the number of channels ($C$) produced at output of each layer (i.e. $\mathcal{O}(C\times C)$), however in DenseNets the number of parameters is proportional to $\mathcal{O}(l_{th}\times k \times k)$ where $l_{th}$ is the layer index and we usually set $k$ much smaller than $C$ so the number of parameters in each layer of the DenseNet is much fewer than that in normal ConvNet.
\item It ensures that there is maximum feature reuse as the features fed to each layer is a consolidation of the features from all the preceding layers and this leads to learning features which are more diversified and pattern rich.

\item DenseNets have shown to be well suited even when the training data is minimal this is because the connectivity pattern in DenseNets ensures that both low and high complexity features are maintained across the network. Hence, the final classification layer uses features from all complexity levels and thereby ensures smooth decision boundaries. Whereas in normal ConvNet the final classification layer builds on top of the last convolution layers which are mostly complex high level features composed of many non-linear transformations.
\end{itemize}

\subsection{Overview of Residual Networks}
\begin{figure}
  \begin{center}
    \includegraphics[width=.4\textwidth]{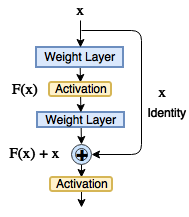}
  \end{center}
  \caption{Residual Learning: A building block}
  \label{fig:resnet}
\end{figure}
Residual Networks (ResNets) are designed to ease the training of very deep networks by introducing a identity mapping (short-cut connections) between input ($x$) and output of a layer ($H(x)$) by performing element-wise summation of the non-linear transformation introduced by the layer and its input. Referring to Figure \ref{fig:resnet}, the residual block is reformulated as $H(x) = F(x)+x$, which consists of the residual function $F(x)$ and input $x$. The idea here is that if the non-linear transformation can approximate the complicated function $H(x)$, then it is possible for it to learn the approximate residual function $F(x)$. 

\subsection{Overview of Inception Architectures}

\begin{figure*}
\centering
\includegraphics[width=1\textwidth]{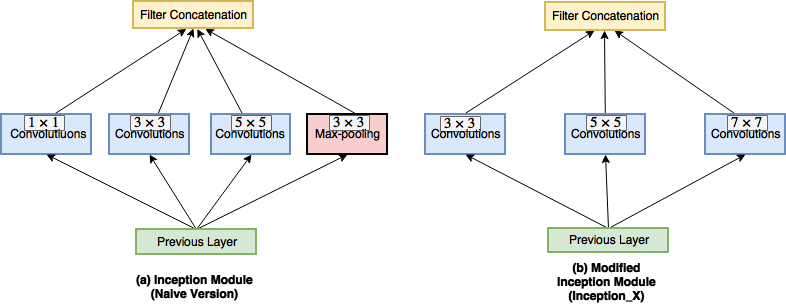}
\caption{Inception module introduces parallel paths with different receptive field sizes by making use of multiple filters with different sizes, e.g. : $1\times 1$, $3\times3$, $5 \times 5$ convolutions and $3\times 3$ max-pooling layer. These feature maps are concatenated at the end. These operations are meant to capture sparse patterns of correlations in the stack of feature maps. The figures: (a) shows the naive version of the inception module, (b) modified version of inception module by excluding max-pooling and introducing a larger kernel ($7\times7$) to increase the receptive field. For the task of semantic segmentation a small kernel helps in detecting small target regions whereas a larger kernel contributes to not only detecting larger target regions but also effectively aids in eliminating false positive regions that have similar properties as the target of interest. 
}
\label{fig:inception_x}
\end{figure*}

The Inception modules were a parallel sub-networks (Figure \ref{fig:inception_x} (a)) introduced in GoogLeNet architecture for ILSVRC 2014 competition, these modules are stacked upon each other, with occasional max-pooling layers with stride 2 to reduce the spatial dimension. The $1\times1$ convolution allowed dimension reduction, thereby making the architecture computationally efficient. These modules were used only in higher layers, whereas the lower layers maintained traditional convolution architecture because of technical constraints (\cite{szegedy2015going}).

For the task of semantic segmentation we proposed to use modified version of the inception module (Figure \ref{fig:inception_x}(b)) only in the first layer, however this could be extended to higher layers. The inception architecture design allows the visual information to be processed at various scales and then aggregated so that the next stage can abstract features from the different scales simultaneously. The ratio of $3\times3: 5\times5 : 7\times7$ convolutions could be skewed (like $2:1:1$) as larger kernels have larger spatial coverage and can capture higher abstractions.   

\section{Cardiac Disease Classification}
\label{app:D}
\subsection{Myocardial Wall Thickness Variation Profile Features}
\begin{figure*}
\centering
\includegraphics[width=1\textwidth]{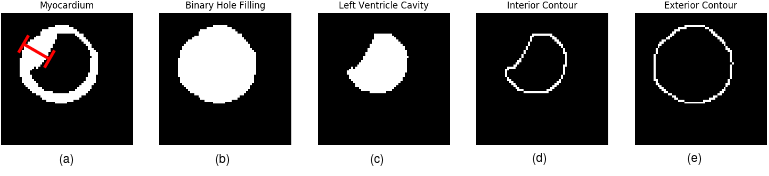}
\caption{The figure illustrates the procedure adopted for estimation of myocardial wall thickness at each short-axis slice. (a) shows the Myocardium segmentation and the red cross-bar indicates the wall thickness at that particular location, (b) shows the binary hole-filling operation on (a), (c) shows the left ventricle cavity got by performing the image subtraction operation between (a) and (b), (d)\& (e) Canny edge detection with $sigma = 1$ was performed on (b) and (c) to get interior and exterior contours.}
\label{fig:myo_thickness}
\end{figure*}
\begin{figure*}
\centering
\subfloat[Confusion Matrix]{\includegraphics[width=.45\textwidth, keepaspectratio]{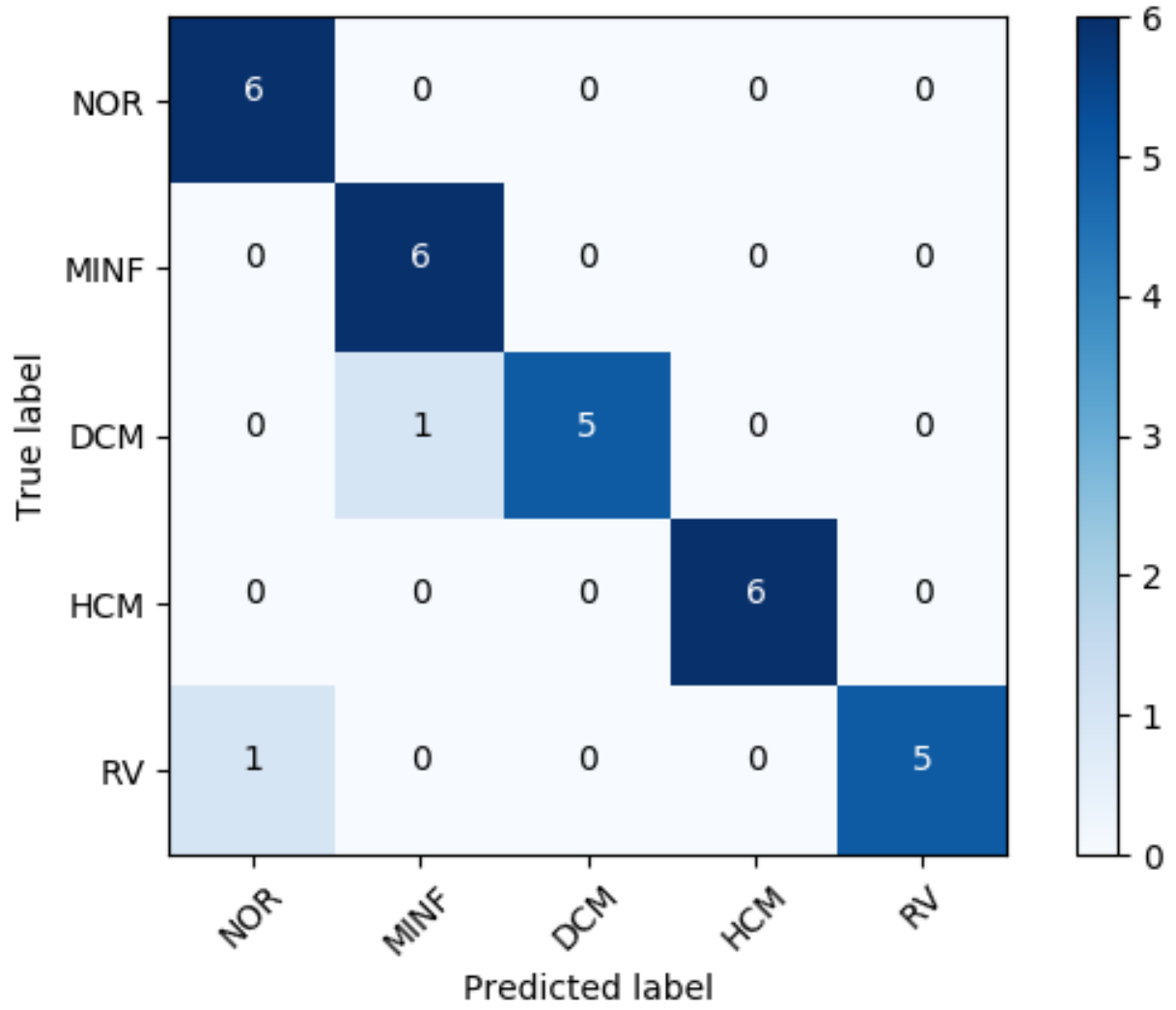}}\hfill
\subfloat[Feature Importance]{\includegraphics[width=0.55\textwidth, keepaspectratio]{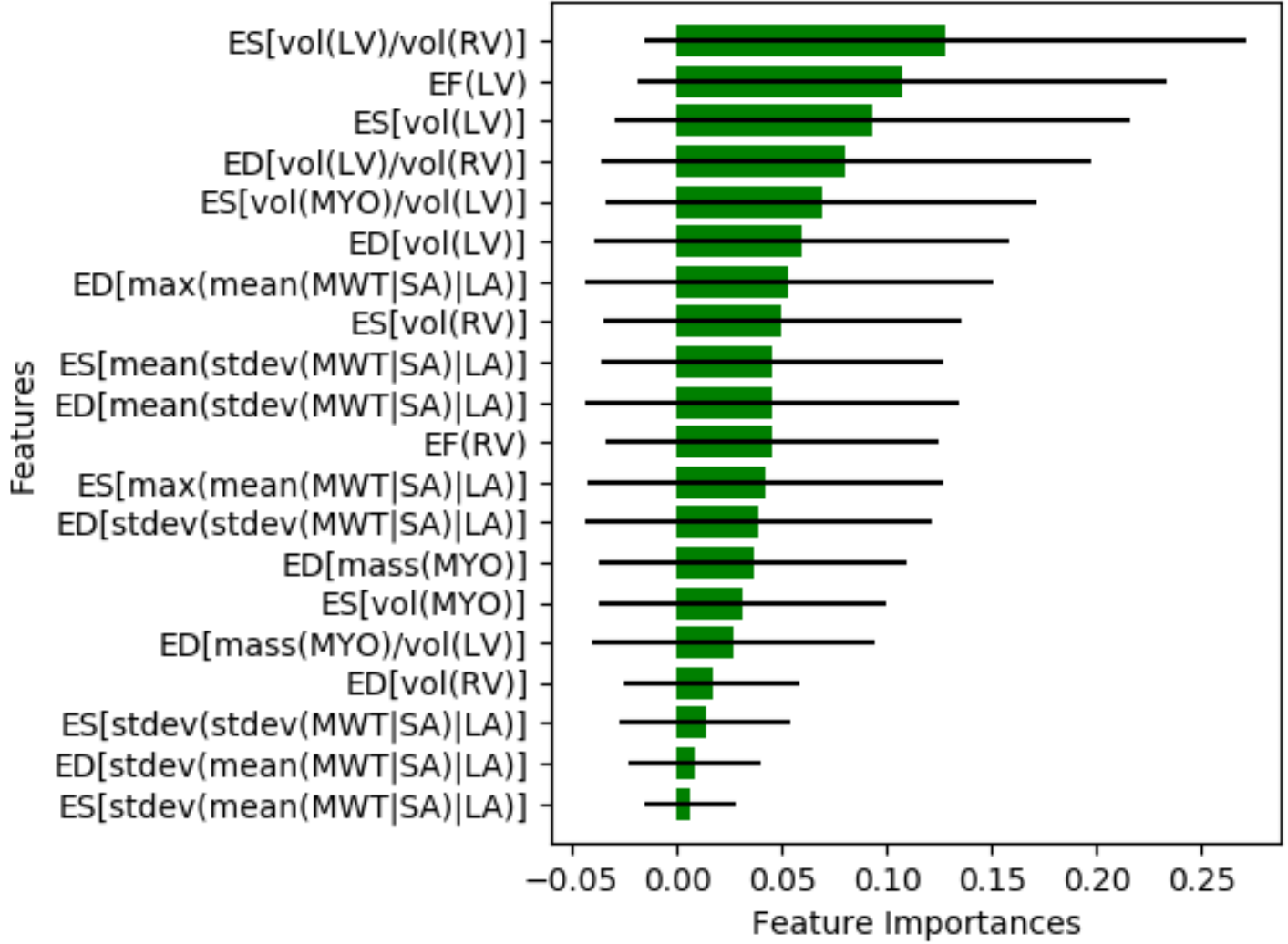}}
\caption{The figure (a) Shows the confusion matrix of the Random Forest predictions on held-out validation set for cardiac disease classification. Rows correspond to the predicted class and columns to the target  class, respectively, (b) Shows the feature importance of all the extracted features by Random Forest classifier. The importance of features on disease classification task was evaluated from individual trees of the Random Forest. The green bars are the feature importances of the forest, along with their inter-trees variability (standard deviation).}
\label{fig:RF}
\end{figure*}

We adopted the following procedure for estimating the myocardial wall thickness variation profile features: 	
\begin{enumerate}
\item The myocardial segmentation mask was subject to canny edge detection to detect the interior and exterior contours. Binary morphological operations like hole-filling and erosion was done to ensure contour thickness to be one pixel width. The Figure \ref{fig:myo_thickness} illustrates the procedure adopted for finding contours.

\item Let $I$ and $E$ be the set of pixels corresponding to the interior $I$ and exterior $E$ contours. Then the Myocardial Wall Thickness (MWT) is the set of shortest euclidean distance (d) measures from a pixel in interior contour $I$ to any pixel in the exterior contour $E$. Formally, the MWT for a Short-Axis (SA) slice is given by:
\begin{equation*}
MWT|SA = \{\min_{e\in E} {d(i,e)}: i \in I\}
\end{equation*}

\item The mean and standard deviation of the MWT is estimated for each SA slice of the heart in ED and ES phases.

\item From the above measurements, $8$ features were derived for quantifying the MWT variation profile at ED and ES phases. The idea was to mathematically quantify how smooth was the variation of average MWT when seen across Long-Axis (LA). Also, how uniform was the MWT in SA slices and to check whether this uniformity was preserved across the slices in LA. The below hand-crafted features were estimated from MWT per SA slices at each cardiac phase:
\begin{itemize}
\item $ES[max(mean(MWT|SA)|LA)]$: Maximum of the mean myocardial wall thickness seen across slices in LA at ES phase.  
\item $ES[stdev(mean(MWT|SA)|LA)]$: Standard deviation of the mean myocardial wall thickness seen across slices in LA at ES phase. 
\item $ES[mean(stdev(MWT|SA)|LA)]$: Mean of the standard deviation of the myocardial wall thickness seen across slices in LA at ES phase.
\item $ES[stdev(stdev(MWT|SA)|LA)]$: Standard deviation of the standard deviation of the myocardial wall thickness seen across slices in LA at ES phase.
\item Similar set of four features were estimated for ED phase.
\end{itemize}
\end{enumerate}
We observed that all classifiers in the first stage of the ensemble had difficulty in distinguishing between DCM vs. MINF disease groups and often misclassified. The Figure \ref{fig:RF} illustrates feature importance ranking of the Random Forest classifier (\cite{liaw2002classification}) and its confusion matrix on validation set. The feature importance of Random Forest classifier suggested that it was giving very low importance to myocardial wall thickness variation profile features and giving more importance to volumetric features like EF. In both MINF \& DCM disease groups the EF of LV is very low and hence could be the reason for confusion. 

\begin{figure*}
\centering
\includegraphics[width=0.8\textwidth]{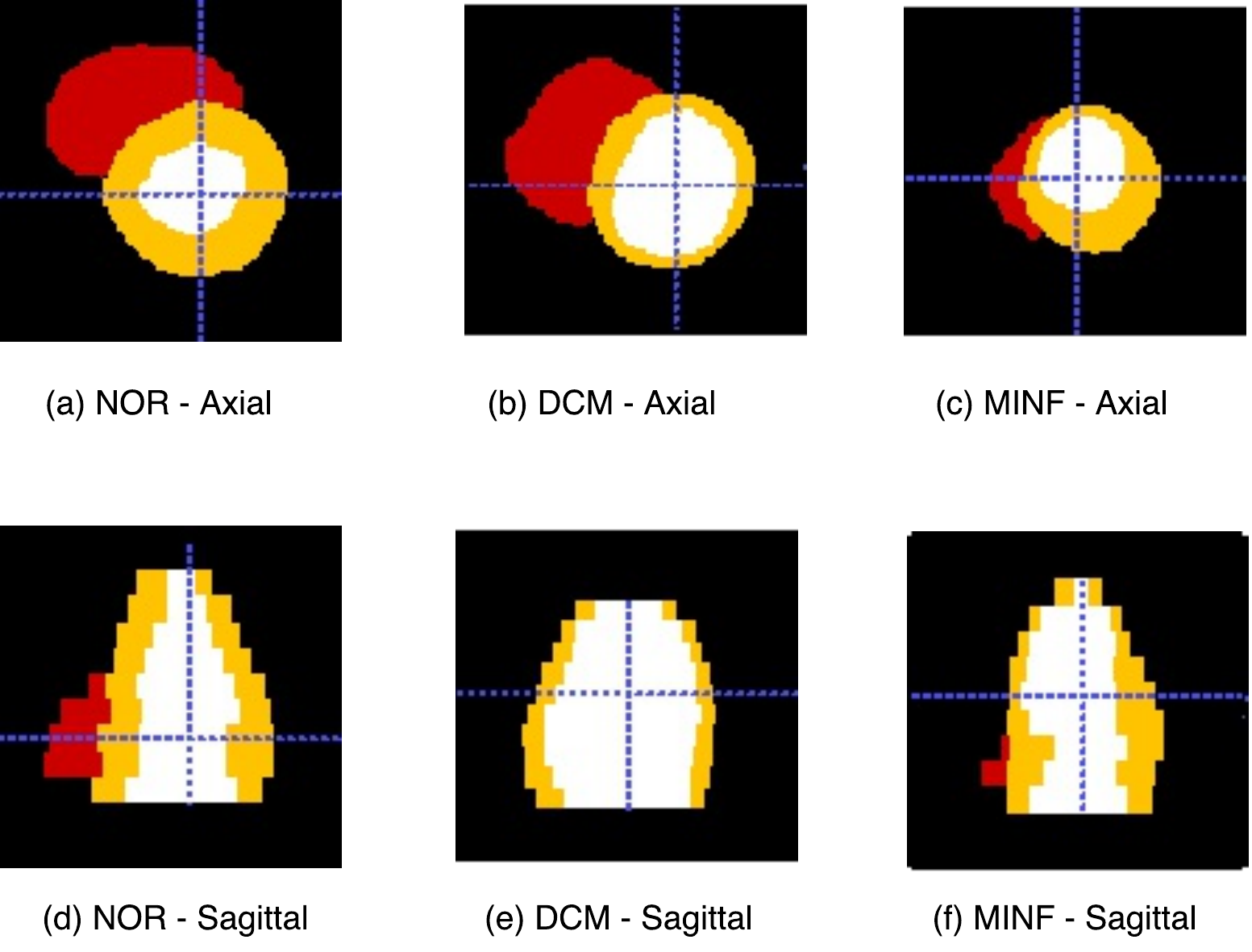}
\caption{The figures shows the short-axis cardiac segmentation of Normal, DCM and MINF cases in axial ((a)-(c)) and sagittal ((d)-(f)) views at End Systole phase. The segmentation labels for RV, LV and MYO are red, yellow and white respectively. For the Normal case the myocardial wall thickness is uniform through out as seen in axial view and its variation along long axis is also smooth. For DCM the myocardial wall is extremely thin when compared to Normal. For MINF, certain sections of the Myocardium wall are extremely thin when compared to rest and hence non-uniformity of thickness is seen and also in long-axis the variation is rough.} 
\label{fig:myo_profile}
\end{figure*}

Clinically, the myocardial wall thickness and its variation profile are the key discriminators in distinguishing between MINF and DCM (\cite{karamitsos2009role}). The myocardial wall thickness peaks during End Systole (ES) phase and is minimal during End Diastole (ED) phase of the cardiac cycle. The myocardial wall thickness changes smoothly in Normal cases. In patients with MINF the wall thickness variation profile is not smooth in both Short-Axis (SA) and Long-Axis (LA) views. Whereas, with DCM cases the wall thickness is extremely thin. Figure \ref{fig:myo_profile} illustrates the myocardial wall thickness variation in Normal, DCM and MINF cases.  

\begin{figure}
\centering
\subfloat[Confusion Matrix]{\includegraphics[width=.4\textwidth, keepaspectratio]{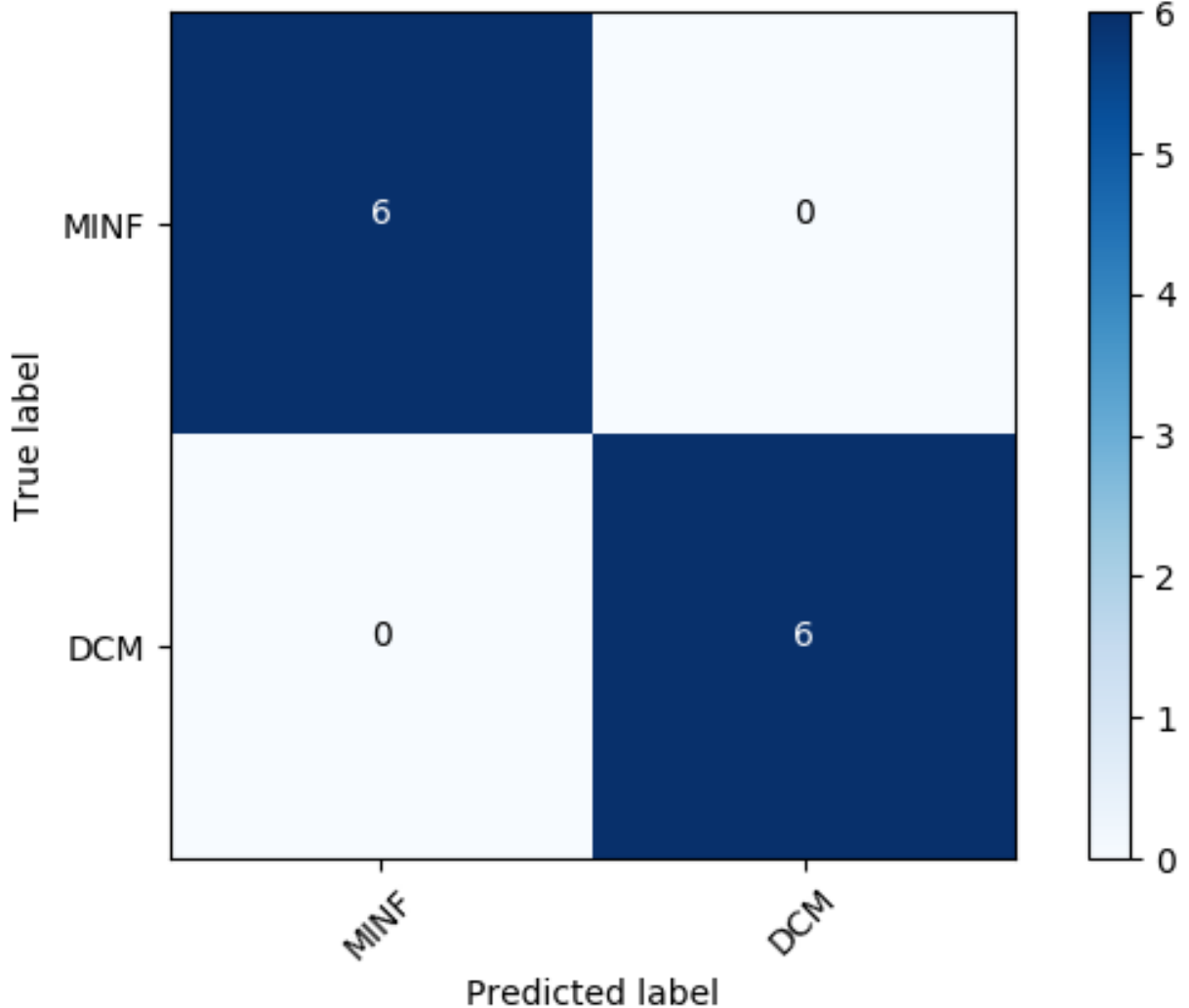}}
\caption{The figure shows the Confusion matrix of the Expert classifier (MLP) predictions on DCM vs. MINF classification task on held-out validation set. Rows correspond to the predicted class and columns to the target class, respectively. The MLP accuracy is 100\%}
\label{fig:cmMLP}
\end{figure}

\pagebreak
\section*{References}

\bibliography{references}

\end{document}